\RequirePackage{fix-cm}
\documentclass[smallextended]{svjour3}       
\smartqed  
\usepackage{graphicx}
\usepackage[font=footnotesize]{caption}
\usepackage{amsmath, amsthm, amssymb, amsfonts}
\usepackage{algorithmic, algorithm}
\usepackage{mathrsfs}
\usepackage{comment}
\usepackage{enumitem}
\usepackage{float,xcolor}
\usepackage{hyperref}
\usepackage{color}
\usepackage[cjk]{kotex}
\usepackage{cite}
\usepackage{xurl}
\usepackage{soul}

\hypersetup{colorlinks,linkcolor={red},citecolor={olive},urlcolor={red}}
\usepackage{subcaption}

\usepackage{url}
\usepackage{lipsum,graphicx,multicol}
\DeclareMathOperator*{\argmin}{arg\,min}

\DeclareMathOperator{\grad}{\nabla}

\usepackage{graphicx}

\def\W{\mathbf{W}}
\def\H{\mathbf{H}}
\def\G{\mathbf{G}}
\def\X{\mathbf{X}}
\def\Y{\mathbf{Y}}
\def\R{\mathbb{R}}

\def\A{\mathbf{A}}

\def\U{\mathbf{U}}

\makeatletter
\renewenvironment{proof}[1][\proofname]{\par
  \pushQED{\qed}%
  \normalfont \topsep6\p@\@plus6\p@\relax
  \trivlist
  \item[\hskip\labelsep
        \itshape
    #1]\ignorespaces
}{%
  \popQED\endtrivlist\@endpefalse
}
\makeatother

\newcommand{\commHL}[1]{{\textcolor{blue}{#1}}}

\renewcommand{\descriptionlabel}[1]{%
	\hspace\labelsep \upshape\bfseries #1%
}

\newcommand*{\affaddr}[1]{#1} 
\newcommand*{\affmark}[1][*]{\textsuperscript{#1}}

\begin{document}

\title{Supervised low-rank semi-nonnegative matrix factorization with frequency regularization for forecasting spatio-temporal data
}

\titlerunning{SSNMF with time/frequency regularization}



\author{Keunsu Kim \protect\affmark[1] \and  Hanbaek Lyu \protect\affmark[2] \and Jinsu Kim \protect\affmark[1] \and Jae-Hun Jung \protect\affmark[1]  
}
\authorrunning{Keunsu Kim et al.} 

\institute{Keunsu Kim \\   \email{keunsu@postech.ac.kr }\\\\
Hanbaek Lyu \\   \email{hlyu@math.wisc.edu}\\\\
Jinsu Kim \\   \email{jinsukim@postech.ac.kr}\\\\
Jae-Hun Jung \\   \email{jung153@postech.ac.kr}\\\\
\affaddr{\affmark[1]The Department of Mathematics, POSTECH, Pohang, Republic of Korea, 37673}\\\\
\affaddr{\affmark[2]The Department of Mathematics, University of Wisconsin–Madison, Wisconsin, USA, 53706}\\
}

\date{Received: date / Accepted: date}

\maketitle

\begin{abstract}
We propose a novel methodology for forecasting spatio-temporal data using supervised semi-nonnegative matrix factorization (SSNMF) with frequency regularization. Matrix factorization is employed to decompose spatio-temporal data into spatial and temporal components. To improve clarity in the temporal patterns, we introduce a nonnegativity  constraint on the time domain along with regularization in the frequency domain. Specifically, regularization in the frequency domain involves selecting features in the frequency space, making an interpretation in the frequency domain more convenient. We propose two methods in the frequency domain: soft and hard regularizations, and provide convergence guarantees to first-order stationary points of the corresponding constrained optimization problem.  While our primary motivation stems from geophysical data analysis based on GRACE (Gravity Recovery and Climate Experiment) data, our methodology has the potential for wider application. Consequently, when applying our methodology to GRACE data, we find that the results with the proposed methodology are comparable to previous research in the field of geophysical sciences but offer clearer interpretability.

\end{abstract}

\section{Introduction}
\label{intro}

Time-series data analysis is primarily divided into two domains: time domain analysis and frequency domain analysis, also referred to as spectral analysis. Spectral analysis specifically examines the periodic properties of time-series data and summarizes them through the spectral density function. The primary objective of spectral analysis is to estimate or infer the spectral density function. In this paper, we propose a method that combines the usual spectral analysis and nonnegative matrix factorization.

Matrix factorization is a technique used to decrease the rank of a matrix by factorizing it into two low-rank matrices and extracting the latent features of the data. A low-rank matrix means that each column (row) is linearly dependent, or in other words, there is a correspondence for each column (row). Among matrix factorizations, nonnegative matrix factorization involves factorizing a nonnegative matrix into two nonnegative matrices. In \cite{lee1999learning}, the nonnegative constraint is used to achieve interpretability. However, in general, datasets may have negative values, so we need to relieve the nonnegative constraint on factorization. Semi-nonnegative matrix factorization is a relaxation technique motivated by K-means clustering, as described in \cite{ding2008convex}.

{\color{black} Matrix factorization is an important technique used in this paper for separating the spatial and temporal components of spatio-temporal data. The concept of breaking down complex problems into multiple components has a long history in mathematical research. For instance, the following statement by Ren\'e Descartes aligns with our statement - 
}
\textit{To divide each of the difficulties under examination into as many parts as possible, and as might be necessary for its adequate solution}.
\footnote{
Discourse on the method of rightly conducting the reason and seeking the truth in the sciences, Ren\'e Descartes (1637) edited by Charles W. Eliot, P.F. Collier \& Son, 1909, New York. 
Transcribed by Andy Blunden. \url{https://www.marxists.org/reference/archive/descartes/1635/discourse-method.htm}

}

By decomposing the data into lower-dimensional matrices, matrix factorization can reveal underlying patterns that are not immediately apparent from the raw data. This approach is similar to the separation of variables in partial differential equations, which is used to simplify the equations and make them easier to solve. In particular, the separation of variables allows us to break down a complex equation into multiple simpler equations, each of which can be solved independently. Matrix factorization also follows the similar decomposition method. Matrix factorization is a unsupervised learning process, but for the prediction problem, we need a supervised learning, particularly in the form of semi-nonnegative matrix factorization. Supervised Nonnegative Matrix Factorization is proposed in \cite{austin2018fully} to solve the classification problem. {\color{black} In \cite{austin2018fully} it is proposed to use regularization in the loss function of nonnegative matrix factorization, which is a cross-entropy term between the ground truth and the class label.} In contrast, our proposed method is a supervised semi-nonnegative matrix factorization with regularization in the frequency domain for forecasting spatio-temporal data. The traditional approach involves employing regularization in the physical domain. However, numerous time-series data characterized by time periodicity,
such as those derived from geophysical applications exhibit distinct characteristics in terms of frequency. In other words, the patterns observed in the physical domain display periodicity aligned with Earth's motion, often with a noticeable yearly frequency. Utilizing such frequency information for regularization yields   convenience in terms of interpretation. Moreover, for forecasting problems, incorporating frequency helps in predicting the spatial patterns that undergo periodic changes. We introduce a novel approach using matrix factorization with regularization in the frequency domain for forecasting problems. 

For the proposed method, we explore two approaches for the proposed regularization method in the frequency domain: soft regularization and hard regularization.  The soft regularization method permits the presence of noisy or insignificant frequencies in the regularization process, whereas the hard regularization method enforces  such frequencies to be eliminated.
Thus, for soft regularization, it is not necessary to pre-specify the frequencies to be removed, but it cannot completely eliminate specific frequencies. We use a high level of block coordinate descent (BCD) and the projected subgradient descent method for the implementation. Further we prove the convergence properties of BCD in this paper. The difficulty arises from the fact that the Fourier transform operates in the complex domain, and the given penalty is not always differentiable. In hard frequency regularization, we apply the results from the three operator splitting method. This algorithm can entirely remove specific frequencies, but it requires prior knowledge to specify which frequencies to remove. To address these challenges with hard frequency regularization, we introduce a heuristic method for solving hard frequency regularization. The heuristic method allows for the removal of specific frequencies without prior knowledge, but there is no guarantee of convergence.

{\color{black}This paper is composed of the following sections. In Section 2 we propose supervised semi-nonnegative matrix factorization with a novel regularization method, that is,  soft/hard regularization in the frequency domain. In Section 3 we present the algorithms to solve the problem proposed in Section 2. In Section 4 we provide theoretical convergence analysis of the algorithms introduced in Section 3. In Section 5 we apply our proposed method  to synthetic data to clarify the differences between the proposed method and existing methods, specifically focusing on matrix factorization and regularization methods. Finally, in Section 6 we apply the proposed method to the GRACE data for forecasting and compare the results with those by the methods currently available in the geophysical community.}

\section{Methods}

\subsection{Problem setup and low-rank spatio-temporal model}
\label{sec:2}

Consider two fixed integer time $T$ and $T_{tot}$, $T_{tot}>T \ge 0$ and suppose that we have tensor data $\mathcal{X}, \mathcal{Y} \in \R^{A\times B \times T_{tot}}$. \label{not:data tensor} We view each of $\mathcal{X}$ and $\mathcal{Y}$ as a `spatio-temporal' data, where $\mathcal{X}[a,b,t]$ encodes the observed value at `spatial location' $(a,b)$ at `time' $t$. One can view $\mathcal{X}$ as the evolution of a real-valued observable (e.g., gravitational force) in a spatial region encoded as a $A\times B$ matrix over the period of discrete time $\{0,\dots,T_{tot}-1\}$, but with the `temporal slices' $\mathcal{X}[:,:, t]$, for times between $T$ and $T_{tot}-1$,  missing. \label{not:T,T_tot} Similarly, we view $\mathcal{Y}$ as the evolution of an auxiliary observable (e.g., precipitation) over the same spatial region over the longer period of time $\{0,\dots,T_{tot}-1\}$. The main question we address in this paper is the following: \textit{Can we find spatio-temporal patterns from the joint tensor data {$(\mathcal{X},\mathcal{Y})$ and use them to predict $\mathcal{X}$ over the missing period of $[T,T_{tot})$?} }

\begin{figure}[h]
\centering
\includegraphics[width=1 \linewidth]{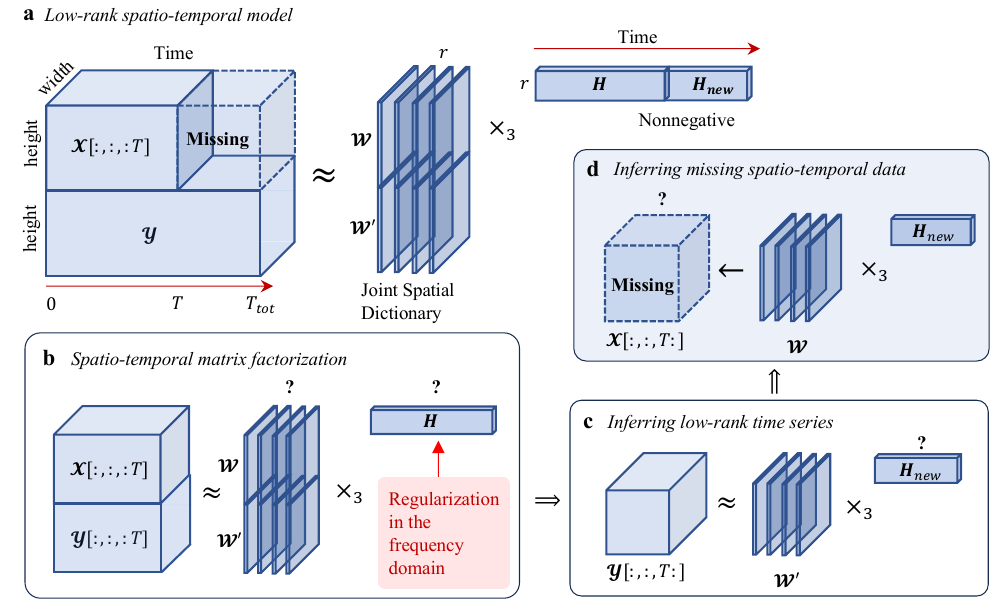}
\caption{(\textbf{a}) Schematic illustration of the low-rank spatio-temporal model. (\textbf{b}) From the joint spatio-temporal data during the period $[0,T)$, we learn the latent spatial patterns $(\mathcal{W},\mathcal{W}')$ and low-rank time-series $\mathbf{H}$, where the temporal structure of $\mathbf{H}$ is regularized in the frequency domain. (\textbf{c}) Using the auxiliary data $\mathcal{Y}$ and the latent dictionary $\mathcal{W}'$, we infer the low-rank time-series $\H_{\textup{new}}$ during the missing period $[T,T_{\textup{tot}})$. (\textbf{d}) The missing data $\mathcal{X}[:,:,T:]$ is then inferred as $\mathcal{W}\times_{3} \H_{\textup{new}}$. 
}
\label{fig:NMF_diagram}
\end{figure}

Our approach is based on the following assumption that \textit{there are $r$ latent spatial patterns and each temporal slice of $\mathcal{X}$ and $\mathcal{Y}$ is their nonnegative linear combination.} Such $r$ \label{not:number of patterns} spatial patterns can be encoded as two tensors $\mathcal{W},\mathcal{W}'\in \R^{A\times B\times r}$. We hypothesize  that each temporal slice $\mathcal{X}[:,:,t]$ and $\mathcal{Y}[:,:,t]$ can be approximated by some linear combination of the $r$ spatial patterns $\mathcal{W}[:,:,s]$ and $\mathcal{W}'[:,:,s]$ for $s=0,\dots,r-1$. That is, there exist \textit{common} and \textit{nonnegative} coefficients $h_{s,t}\ge 0$ for $s=0,\dots,r-1$ such that 
\begin{align}\label{eq:model_1}
    \mathcal{X}[:,:,t]\approx \sum_{s=0}^{r-1} \mathcal{W}[:,:,s] h_{s,t}, \quad \mathcal{Y}[:,:,t]\approx \sum_{s=0}^{r-1} \mathcal{W}'[:,:,s] h_{s,t}.  
\end{align}
Hence the pair of $(\mathcal{W}[:,:,s], \mathcal{W}'[:,:,s])$ encodes possible spatial patterns of the two observables  recorded in $\mathcal{X},\mathcal{Y}$ that can be \textit{simultaneously observed} in the spatial region $A\times B$. The underlying latent statio-temporal model can be concisely re-stated as the following tensor factorization equations:
\begin{align}\label{eq:low_rank_model}
    \mathcal{X} \approx \mathcal{W} \times_{3} \widetilde{\H}, \quad \mathcal{Y}\approx \mathcal{W}' \times_{3} \widetilde{\H},
\end{align}
where $\times_{3}$ denotes the mode-3 tensor-matrix product and $\H\in \R^{r\times T}_{\ge 0}$ \label{not:temporal pattern} and $\widetilde{\H}[s,t]=h_{s,t}$ for $t \in [0,T_{tot})$ encode the rank-$r$ representation of the joint time-series $(\mathcal{X},\mathcal{Y})$ over the basis of $(\mathcal{W},\mathcal{W}')$. We call \eqref{eq:low_rank_model} the \textit{low-rank spatio-temporal model}
(See Fig.  \ref{fig:NMF_diagram}\textbf{a} for an illustration.) 

\begin{remark}
\textit{We remark that extending our model for multiple auxiliary spatio-temporal data $\mathcal{Y}_{0},\dots,\mathcal{Y}_{S-1}\in \R^{A\times B\times T_{tot}}$ \label{not:S} is straightforward by simply concatenating them into a single $AS\times B \times T_{tot}$ tensor $\mathcal{Y}$ and use the same framework as in \eqref{eq:low_rank_model} except that the `height' of $\mathcal{Y}$ and $\mathcal{W}'$ is now $AS$ instead of $A$. For the simplicity of the discussion, we keep the setting $S=1$. But later in our experiments, we use an instance of $S=3$ where $\mathcal{Y}_0$ for normalized precipitation ranging between $0$ and $1$, $\mathcal{Y}_1$ for normalized temperature, and $\mathcal{Y}_2$ for normalized  total water storage from the GLDAS Noah model.}
\end{remark}

\subsection{Infering the missing spatio-temporal data given the latent spatial patterns}

The unknown parameters for the low-rank spatio-temporal model \eqref{eq:low_rank_model} are the latent spatial patterns $(\mathcal{W}, \mathcal{W}')\in \R^{2A \times B \times r}$ and the low-rank nonnegative time-series $\widetilde{\H}\in \R_{\ge 0}^{r\times T_{tot}}$. Learning these parameters simultaneously should be carefully formulated through constrained nonconvex optimization problems, which we will discuss in the following section. Here we first discuss how we can infer the missing spatio-temporal data $\mathcal{X}[:,:,T\mathord:]\in \R^{A\times B\times (T_{\textup{tot}}-T)}$ by assuming that we have the latent spatial patterns  $(\mathcal{W}, \mathcal{W}')$. Here the notation $T\mathord:$ means all the elements starting from index $T$ to the end. In a similar way, 
$:\mathord T$ means all the elements until index $T-1$  excluding $T$. This can be done through the following three steps as below:
\begin{align}
\text{(Encoding)} \quad   \H_{\textup{new}}' & \leftarrow \argmin_{\U\in \R_{\ge 0}^{r\times T_{\textup{tot}}}} \lVert \mathcal{Y} - \mathcal{W}'\times_{3} \U   \rVert_{F}^{2} \label{eq:inffer_H_new} \\
\text{(Slicing)} \quad \H_{\textup{new}} & \leftarrow \H_{\textup{new}}'[:,T\mathord:] \\
\nonumber \\
 \text{(Prediction)} \quad  \mathcal{X}[:,:,T\mathord:] &\leftarrow \mathcal{W} \times_{3}  \H_{\textup{new}}, \label{eq:infer_X_missing} 
\end{align}
where the subscript $F$ denotes the Frobenius norm.

Namely, according to the low-rank spatio-temporal model \eqref{eq:low_rank_model}, the auxiliary data $\mathcal{Y}$ during the missing period $[T,T_{\textup{tot}})$ is represented by the linear model $\mathcal{Y}\approx \mathcal{W}'\times_{3} \H_{\textup{new}}$ for some nonnegative matrix $\H_{\textup{new}}$. Hence, by solving the nonnegative least-squares problem in \eqref{eq:inffer_H_new}, we can infer the low-rank time-series $\H_{\textup{new}}$. Now $\H_{\textup{new}}$ is also used to represent the missing data $\mathcal{X}[:,:,T\mathord:]$ over the corresponding latent spatial patterns in $\mathcal{W}$. Thus, we can infer the missing spatio-temporal data as in \eqref{eq:infer_X_missing}. See Fig. \ref{fig:NMF_diagram}\textbf{c}-\textbf{d} for an illustration. 

\begin{remark}
\label{remark:loss of frequency}
\textit{It is crucial to construct $\H_{\textup{new}}$ by considering the entire period of $\mathcal{Y}$ in \eqref{eq:inffer_H_new} to accurately infer the missing data. While it may seem doable to use $\mathcal{Y}[:,:,T\mathord:]$ for this purpose, it is important to note that in the frequency domain, only $\lfloor {T \over 2} \rfloor + 1$ frequencies are present for the time-series of length $T$. Consequently, relying solely on the test period information $[T, T_{tot})$ of $\mathcal{Y}$ to construct $\H_{\textup{new}}$ could result in a loss of frequency information. For instance, when the testing period spans 10 months, Encoding step of  \eqref{eq:inffer_H_new} solely with this period would fail to capture the annual patterns. However, by considering the entire duration, even within those 10 months, we can infer the annual patterns reasonably.}
\end{remark}

\subsection{Spatio-temporal matrix factorization with frequency regularization}
\label{sec:OPT_SSNMF}

In this section, we introduce and propose the matrix factorization method with frequency regularization. 
In general, due to the nature of time-series data, specific frequencies often hold significant physical meanings. Thus, analysis with these frequencies can greatly facilitate the meaningful interpretation of the underlying physics in the given data. Consequently, regularization in the frequency domain may be more advantageous  for matrix factorization than regularizing in the time  domain. In this section, we aim to propose methods that incorporate this idea.

Decompose the low-rank time-series $\widetilde{\H}$ as $[\H, \H_{\textup{new}}]$\label{not:tilde H}, where $\H\in \R_{\ge 0}^{r\times T}$ and $\H_{\textup{new}} \in \R_{\ge 0}^{r\times T_{tot} - T}$. With a slight abuse of notation, we will denote $\mathcal{X}[:,:,:T]$ and $\mathcal{Y}[:,:,:T]$ as $\mathcal{X}$ and  $\mathcal{Y}$, respectively, in this section. The main task of this method is to learn the joint latent spatial patterns $(\mathcal{W},\mathcal{W}')$ and the common nonnegative low-rank time-series $\H$ from the joint observed time-series data $(\mathcal{X}, \mathcal{Y})\in \R^{2A\times B\times T}$. 
Here, we make the assumption that the observations were gathered over a sufficient period of time, and we consider the loss of frequency information noted in Remark \ref{remark:loss of frequency} is insignificant. The necessary duration for sufficient observation time can vary depending on the specific data and the temporal pattern under consideration. For instance, for  the GRACE data,  as discussed in Section \ref{sec:Experimental results}, a minimum of $12$ months of observations is required to discern the expected annual pattern. With this assumption
we formulate the problem as the following \textit{spatio-temporal matrix factorization} problem over $[0, T)$: 
\begin{align}\label{eq:SSNMF_learning1}
    \textup{minimize} & \quad \lVert \mathcal{X} - \mathcal{W} \times_{3} \H \rVert_{F}^{2} + \xi  \lVert \mathcal{Y} - \mathcal{W}' \times_{3} \H \rVert_{F}^{2}  + \psi(\H), \\
    \textup{subject to} & \quad \text{$\mathcal{W}, \mathcal{W}' \in \R^{A\times B \times r}$\text{ and } $\H\in \R^{r\times T}_{\ge 0}$ } \nonumber
\end{align}
where $\xi \ge 0$ is a \textit{supervision parameter} and $\psi(\H)$ is a penalty term for $\H$, which will take one of the following choices below:
\begin{description}
    \item[(a)] (Ridge penalty) $\psi(\H) = \lambda \lVert \H \rVert_{F}^{2}$ 
    \item[(b)] (Lasso penalty) $\psi(\H) = \lambda \lVert \H \rVert_{1}$ 
    \item[(c)] (Soft frequency regularization) $\psi(\H) = \lambda \lVert \widehat{\H} \rVert_{1,M}$ \quad ($\triangleright$ $\lVert \cdot\rVert_{1,M}=$Minkowski 1-norm) 
    \item[(d)] (Hard frequency regularization) $\psi(\H) = 0$ if $\widehat{\H}$ only uses a prespecified set of frequencies and $\infty$ otherwise.
\end{description}
Here $\lambda\ge 0$ is a regularization parameter and $\widehat{\H}$ denotes the Fourier transform of $\H$.

The rationale behind regularizing the temporal structure of $\H$ in  \eqref{eq:SSNMF_learning1} is as follows. The observed data ($\mathcal{X}$ as well as $\mathcal{Y}$) may contain noise so we should avoid overfitting the latent spatial patterns to accurately represent the observation. Instead of directly regularizing the latent spatial patterns, we find it more effective to regularize the associated low-rank time-series representation $\H$ so that it is \textit{sparse in the frequency domain}. That is, our guiding principle is that the observed spatio-temporal data admits a low-rank time-series representation $\H$, which uses a small number of dominating frequencies. Our intuition is largely inspired by analyzing geospatial data (e.g., GRACE data, see Sec. \ref{sec:GRACE}), where typically the annual or semi-annual spatio-temporal patterns are dominating, and other frequencies are often under-represented. Options (\textbf{a}-\textbf{b}) above use standard Ridge and Lasso (least absolute shrinkage and selection operator) penalties commonly used in the machine learning and statistics literature \cite{hastie2009elements}. (We opt for using the Lasso penalty being a convex relaxation of the computationally more challenging $\ell_{0}$-penalty \cite{ramirez2013l1}.) Note that both penalties \textit{regularize $\H$ in the time domain}, whereas the options (\textbf{c}-\textbf{d}) regularize the Fourier transform $\widehat{\H}$ of $\H$ so the regularization is done directly in the frequency domain. We will discuss the latter two options in more detail in the next section.

Notice that we referred to \eqref{eq:SSNMF_learning1} as a matrix factorization problem while it involves factorizing tensors $\mathcal{X}$ and $\mathcal{Y}$. This is because one can matricize these tensors as well as the latent spatial patterns $(\mathcal{W},\mathcal{W}')$ to reformulate \eqref{eq:SSNMF_learning1} as a matrix factorization problem. Namely, let $\X:=(\mathcal{X}_{(3)})^{T}\in \R^{AB\times T}$ \label{not:matricization} denote the matrix by matricizing the tensor $\mathcal{X}$ along the time mode, where $\mathcal{X}_{(3)}\in \R^{T\times AB}$ denotes the mode-3 unfolding of $\mathcal{X}$. Similarly, denote $\Y:=(\mathcal{Y}_{(3)})^{T}$, $\W:=(\mathcal{W}_{(3)})^{T}$, and $\W':=(\mathcal{W}'_{(3)})^{T}$. \label{not:spatial pattern} Then we have $\lVert \mathcal{X} - \mathcal{W} \times_{3} \H \rVert_{F} = \lVert \X - \W  \H \rVert_{F} $ and $\lVert \mathcal{Y} - \mathcal{W}' \times_{3} \H \rVert_{F} = \lVert \Y - \W'  \H \rVert_{F} $. So henceforth, we will consider the following \textit{supervised semi-nonnegative matrix factorization} (SSNMF) problem instead of \eqref{eq:SSNMF_learning1}: (denoting $d:=AB$)
\begin{align}\label{eq:SSNMF_main}
    \textup{minimize} & \quad \lVert \X - \W \H \rVert_{F}^{2} + \xi  \lVert \Y - \W' \H \rVert_{F}^{2}  + \psi(\H) \\
    \textup{subject to} & \quad \text{$\W, \W' \in \R^{d \times r}$\text{ and } $\H\in \R^{r\times T}_{\ge 0}$ }. \nonumber
\end{align}

\section{Algorithms}
\label{sec:algorithms}

\subsection{Soft frequency regularization}

{
Spectral analysis is popularly used in time-series analysis. It uncovers the underlying periodic patterns present in the given time-series data. This method revolves around the concept of the spectral density function, which is a probability density function describing the distribution of the modulus of frequencies across different frequencies in the data. The primary goal of spectral analysis is to accurately estimate the spectral density function using the sampled time-series data. The discrete Fourier transform is widely utilized for such estimation. {\color{black}However, the expectation value of the periodogram at a certain frequency 
does not converge to the spectral density function at that frequency 
even when dealing with a large number of samples.}} To mitigate this issue, smoothing in the frequency domain is commonly employed in practical time-series analysis \cite{brockwell2016introduction}. We aim to integrate spectral analysis in classical time-series data analysis with SSNMF. That is, instead of analyzing the time-series data corresponding to each spatial data, we utilize  regularization from the perspective of analyzing in the frequency domain of the time-series data obtained through the dimensional reduction of the given spatio-temporal data using SSNMF. In Proposition \ref{prop:synthetic_code_closed_form}, a mismatch term between $\widehat{\X}$ and $\widehat{\Y}$ can disrupt the inference of the periodic pattern of $\X$. Therefore, regularization in the frequency domain may help mitigate the impact of this disparity in forecasting.

\begin{definition}[Fourier transform]
\label{not:Fourier transform}
For a matrix $\A\in \R^{r\times T}$, we regard its rows $\A[s]$ for $s=0,\dots,r-1$ as a univariate time-series where the column index corresponds to time. With this interpretation, we can define the \textit{Fourier transform} of $\A$, denoted as $\widehat{\A}$, to be the $r\times T$ matrix of complex coefficients defined as 
\begin{align}
        \widehat{\A}[s,k] := {1\over T}\sum\limits_{f=0}^{T-1}\A[s,f]e^{-2\pi i f k /T}
\end{align}
for $0\le s \le r-1$, $0\le k \le T-1$. Denote by $\mathcal{F}_{T}$ the $T\times T$ \textit{Fourier matrix} whose $(f,k)$ coordinate equals $e^{-2\pi i f k /T}/T$. Then we can write 
\begin{align}
    \widehat{\A}  = \A \mathcal{F}_{T}. 
\end{align}    
\end{definition}

In the case of Lasso, when we reduce the number of features, it simplifies the interpretation of each parameter. Similarly, when we introduce sparsity in the frequency domain, it simplifies the interpretation of the patterns in the temporal data obtained through matrix factorization. 
Therefore, we propose the following statio-temporal decomposition problem with frequency regularization: 

\begin{equation}
\label{eq:loss function}
    \min_{\substack{\W,\W' \in \R^{d\times r} \\ \widehat{\H} \in \mathbb{C}^{r\times T}, \\ \widehat{\H}_{sk} = \overline{\widehat{\H}}_{s,T-k},  \widehat{\H}\mathcal{F}_T^{-1} \ge 0}}  \lVert \widehat{\X}[:,:T]-\W\widehat{\H} \rVert_{F}^{2}  + \xi \lVert \widehat{\Y}[:,:T]-\W'\widehat{\H} \rVert_{F}^{2} + \lambda \lVert \widehat{\H} \rVert_{1,M}.\end{equation}
Here, $\xi,\lambda\ge 0$ are tuning parameters and for a complex matrix $C\in \mathbb{C}^{r\times T}$, $\lVert C \rVert_{1,M}$ denotes its \textit{Minkowski 1-norm} defined as 
\begin{align}
\label{not:Minkowski norm}
    \lVert C \rVert_{1,M} := \sum_{a,b}  \lVert C[a,b] \rVert_{1,M}, \,\, \lVert x+i y \rVert_{1,M} := |x| + |y|. 
\end{align}
The condition $\widehat{\H}_{sk} = \overline{\widehat{\H}}_{s,T-k}$ is needed because we want $\H$ to be a real matrix and the condition $\widehat{\H}\mathcal{F}_T^{-1}\ge 0$ is applied to the nonnegative constraint on $\H$.

Note that by the Parseval's identity, $\lVert \widehat{\X} - \W\widehat{\H} \rVert_F^2 =\lVert (\X - \W\H)\mathcal{F}_T \rVert_F^2 = \lVert \X - \W\H \rVert_F^2$.   
Therefore problem \eqref{eq:loss function} is equivalent to
\begin{equation}
\label{eq:loss function2}
\min_{\substack{\H\in \mathbb{R}_{\ge 0}^{r\times T} \\ \W, \W'\in \mathbb{R}^{d\times r}}} \lVert \X - \W\H \rVert_F^2 + \xi \lVert \Y - \W'\H \rVert_F^2 + \lambda \lVert \H \mathcal{F}_T \rVert_{1,M}.  
\end{equation}
Moreover, $\lVert \widehat{\H} \rVert_{1,M} = {1 \over T} \lVert \H \rVert_{1} + \sum\limits_{s,k=1} \lVert \widehat{\H}[s,k] \rVert_{1,M}$ holds. This relation tells us that the Minkowski $1$-norm contains $L1$-regularization term.

Algorithm \ref{alg:main} below takes the form of block coordinate descent for solving SSNMF \eqref{eq:SSNMF_main} for three instances of regularization: Ridge and Lasso regularization in the time domain; and soft regularization in the frequency domain (see \eqref{eq:loss function2}).

\begin{algorithm}[H]
\begin{algorithmic}[1]
\caption{SSNMF with Ridge, Lasso and soft frequency regularization}
\label{alg:main}

\STATE \textbf{Input:} $\X\in \R^{d\times T}$ (Data) and $\Y \in \R^{d \times T_{tot}}$ (Auxiliary data);
\vspace{0.1cm}
\STATE \textbf{Variables:} $N\in \mathbb{N}$ (iterations), $L \in \mathbb{N}$ (sub-iterations);\,  $\lambda  \ge 0$ (regularization parameter);
\STATE \textbf{Regularizer:} $\psi(\H)=\lVert  \H\rVert_{F}^{2}$ (Ridge) or $\lVert \H \rVert_{1}$ (Lasso) or $\lVert \H \mathcal{F}_T \rVert_{1,M}$ (soft frequency regularization)
\STATE \textit{Initialize:} \quad  $\mathbf{W}_{0}, {\mathbf{W}'}_{0} \in \mathbb{R}^{d\times r }$ and $\mathbf{H}_{0}\in \mathbb{R}^{r\times T}$

\vspace{0.1cm}
\FOR{$j = 1,\cdots ,N$}
\STATE
\begin{align}
\hspace{-0.5cm} \label{eq:coding_H} \H_j &\leftarrow \argmin\limits_{\H\in \mathbb{R}_{\ge 0}^{r\times T}} \left \lVert \begin{bmatrix} \X  \\ \sqrt{\xi}\Y[:,:T] \end{bmatrix} - \begin{bmatrix} \W_{j-1}  \\ \sqrt{\xi}{\W'}_{j-1} \end{bmatrix} \H \right\rVert_F^2  + \lambda \psi(\H)  \\ 
&\qquad (\textup{$\triangleright$ Use projected subgradient descent. Sub-iteration $L$ is required here.}) \nonumber \\
\W_j &\leftarrow \argmin\limits_{\W\in \mathbb{R}^{d\times r}} || \X  - \W \H_j||_F^2 \quad \textup{$\left( \triangleright \,\,  \W_j = \X {\H_j}^T (\H_j{\H_j}^T)^{-1} \right)$} \label{eq:coding_W0} \\
\hspace{-0.5cm} \W'_j &\leftarrow \argmin\limits_{\W'\in \mathbb{R}^{d \times r}} || \Y[:,:T]  - \W' \H_j||_F^2 \quad \left( \triangleright \,\, {\W'}_j = \Y[:,:T] {\H_j}^T (\H_j{\H_j}^T)^{-1}  \right) \label{eq:coding_W1}
\end{align}
\ENDFOR

\vspace{0.1cm}
\STATE (Encoding) ${}$
\vspace{0cm} 
\begin{align}
\label{eq:encoding_H}
\hspace{-0.5cm}\H_{\textup{new}}' &\leftarrow \argmin\limits_{\H\in \mathbb{R}_{\ge 0}^{r\times T_{tot}}} || \Y  - \W' \H||_F^2 + {\lambda \over \xi} \psi(\H) \\
& (\textup{$\triangleright$ Use projected subgradient descent. Sub-iteration $L$ is required here.}) \nonumber \\
\H_{\textup{new}}&\leftarrow \H'_{\textup{new}}[:,T:] 
\end{align}

\vspace{0.1cm}
\RETURN{} $\W_{N}, \W'_{N}, \H_{N}, \H_{\textup{new}}$.

\end{algorithmic}
\end{algorithm}

\subsection{Hard constraints in the frequency domain}
\label{sec:Hard constraints}
{\color{black} 
Lasso introduces sparsity in the time domain, allowing us to select proper features. Similarly, we employ soft frequency regularization to introduce sparsity in the frequency domain. However, in our experiments (Examples \ref{ex:three norms} and \ref{ex:Lasso soft grace}), we observed that the soft constraint reduces the utilization of less dominant frequencies but does not  completely eliminate them. As a result, the soft constraint in the frequency domain partially fulfills our objective. This partial achievement led us to introduce the hard constraint in the frequency domain, enabling us to selectively eliminate specific frequencies.
} 
Our algorithm is based on the three operator splitting problem, and we have proven its convergence. {\color{black}However, for this algorithm 
specific frequencies that need to be eliminated should be predetermined in advance.
} And determining these frequencies necessitates prior knowledge of temporal patterns. Therefore, we introduce a heuristic method to address the hard constraint problem, which does not rely on prior knowledge. By doing so, we can achieve a clear temporal pattern related to the corresponding spatial information. This additional step enhances our ability to identify and interpret the meaningful patterns in the data.

Consider the three operator splitting problem to find $x \in \mathbb{R}^{n}$ such that 

\begin{equation}
\min_{x \in \mathbb{R}^{n}} f(x)+g(x)+h(x).
\end{equation}

If $g$ and $h$ are indicator functions on the convex domain $\mathcal{G}$ and $\mathcal{H}$, respectively, then the three operator splitting problem becomes 
\begin{equation}
\label{eq:three operator}
\min_{x \in \mathbb{R}^{n}} f(x) \text{ subject to }x \in \mathcal{G} \cap \mathcal{H}.
\end{equation}

The method for finding a value of $x$ is referred to as the following proposition. Define the ergodic sequences $\bar{x}_t = {1 \over t+1} \sum\limits_{\tau =0}^t x_{\tau}$. To prioritize satisfying the nonnegative condition over suppressing specific frequencies, we set
\begin{align}
f(\H) &= \lVert \X - \W \H \rVert_{F}^{2} + \xi \lVert \Y - \W' \H \rVert_{F}^{2}, \label{eq:f}\\
\nonumber
\\
g(\H) &= \begin{cases} 0, & \mbox{if } \H\ge 0 \\ \infty, & \mbox{ Otherwise} \end{cases}, \label{eq:g}\\
\nonumber
\\
h_{R}(\H) &= \begin{cases} 0, & \mbox{if } \H \in \mathcal{F}_{T}^{-1} (\mathbb{C}^{r \times R} \times \left\{ 0 \right\}^{T-R} ) \label{eq:h}\\ \infty, & \mbox{ Otherwise} \end{cases},
\end{align} here, the subscript $R$ represents the number of frequencies to remain. If the priority is to suppress specific frequencies rather than to satisfy the nonnegative condition, the functions $g$ and $h_R$ should be interchanged.

\begin{algorithm}[H]
\begin{algorithmic}[1]
\caption{Three Operator Splitting \cite{yurtsever2021three}}
\label{alg:TOS}

\STATE \textbf{Input:} Initial point $y_0 \in\mathbb{R}^{r\times T}$; $R \in \mathbb{N}$ (the number of remaining frequencies)
\vspace{0.1cm}

\vspace{0.1cm}
\FOR{$j = 0,\cdots ,N$}
\STATE
\begin{align}
\hspace{-0.5cm}    \H_j = \argmin_{\H \in \R^{r\times T}}\left\{{\gamma_j g}(\H) + {1 \over 2}\lVert y_j - \H \rVert^2 \right\} \\
\hspace{-0.5cm}    
\G_j = \argmin_{\G \in \R^{r\times T}}\left\{{\gamma_j h_R}(\G) + {1 \over 2}\lVert 2\H_j - y_j  - \gamma_j \nabla f(\H_{j}) - \G \rVert^2 \right\} \\
\hspace{-0.5cm}    y_{j+1} = y_j - \H_j + \G_j \\
\hspace{-0.5cm} \gamma_j = {1 \over \sqrt{\sum\limits_{\tau =0}^{j-1} \lVert \nabla f(\H_{\tau}) \rVert^2}}
\end{align}

\ENDFOR
\RETURN{} Ergodic sequence $\overline{\H}_N$.

\end{algorithmic}
\end{algorithm}

\begin{algorithm}[H]
\begin{algorithmic}[1]
\caption{SSNMF with hard constraint in the frequency domain}
\label{alg:Hard2}

\STATE \textbf{Input:} $\X\in \R^{d \times T}$ (Data) and $\Y\in \R^{d \times T_{tot}}$ (Auxiliary data); 
\vspace{0.1cm}
\STATE \textbf{Variables:} $N\in \mathbb{N}$ (iterations);\,  $\lambda  \ge 0$ (regularization parameter); $R \in \mathbb{N}$ (the number of remaining frequencies);
\STATE \textit{Initialize:} \quad  $\mathbf{W}_{0}\in \mathbb{R}^{d \times r }$ and $\mathbf{H}_{0}\in \mathbb{R}^{d \times T}$

\vspace{0.1cm}
\FOR{$j = 1,\cdots ,N$}
\STATE
${}$\vspace{-0.5cm}
\begin{align}
\hspace{-0.5cm} \H_j &\leftarrow \text{Algorithm \ref{alg:TOS} or Algorithm \ref{alg:Heuristic}} \label{alg:hard sub}\\
\hspace{-0.5cm} \label{eq:coding_W0 hard}\W_j &\leftarrow \argmin\limits_{\W\in \mathbb{R}^{d\times r}} || \X  - \W \H_j||_F^2 \quad \textup{$\left( \triangleright \text{ i.e. } \W_j = \X \H_j^T (\H_j\H_j^T)^{-1} \right)$}\\
\hspace{-0.5cm} \label{eq:coding_W1 hard}\W'_j &\leftarrow \argmin\limits_{\W'\in \mathbb{R}^{d\times r}} || \Y[:,:T]  - \W' \H_j||_F^2 \quad \textup{$\left( \triangleright \text{ i.e. } \W'_j = \Y[:,:T] \H_j^T (\H_j\H_j^T)^{-1}\right)$}
\end{align}

\ENDFOR

\STATE ${}$ \vspace{-0.5cm}
\begin{align}
\hspace{-0.5cm}\H_{\textup{new}}' &\leftarrow \text{Algorithm \ref{alg:TOS} or Algorithm \ref{alg:Heuristic}} \label{alg:hard sub_encoding} \\
&\text{for } f(\H) = \lVert \Y - \W'_{\commHL{N}} \H \rVert_{F}^{2} \\
\H_{\textup{new}}&\leftarrow \H'_{\textup{new}}[:,T:] 
\end{align}

\RETURN{} $\W_{N}, \W'_{N}, \H_{N}, \H_{\textup{new}}$.

\end{algorithmic}
\end{algorithm}

Instead of Algorithm \ref{alg:TOS}, we found experimentally that the following algorithm also {\color{black} converges} for updating $\H$, even though there is no convergence guarantee. Moreover, unlike the previous method, the advantage of this method is that it does not require specifying the frequencies to be removed in advance. Furthermore, Algorithm \ref{alg:TOS} should also be carefully applied during the encoding process. In Example \ref{ex : Fourier rescaling}, {\color{black} we illustrate an example of time-series data with the same period but different 
lengths. If we choose Algorithm \ref{alg:TOS} in Algorithm \ref{alg:Hard2} (in the step of (\ref{alg:hard sub})), we should consider the difference of the time-series length when obtaining $\H \in \R^{r \times T}$ and $\H_{\textup{new}}' 
\in \R^{r \times T_{tot}}$.} If we remove specific frequencies during the coding process \eqref{alg:hard sub}, how can we remove frequencies during the encoding process \eqref{alg:hard sub_encoding}? Assume that the training period is $T = 132$ months, and the total period (including both the training and test periods) is $T_{tot} = 163$ months. For example, in coding process the $11$th Fourier coefficient corresponds to a $12$ month cycle but, during the encoding process, the $11$th Fourier coefficient corresponds to the 14.8 month cycle. This raises the question of how to handle frequency removal in the encoding process.

\begin{example}
\label{ex : Fourier rescaling}
For $M = 2T$, let $f(t) = \cos{2\pi t \over T}$ for $t=0, \dots, T-1$ and $F(t) = \cos{4\pi t \over M}$ for $t=0, \dots, M-1$, i.e. $f$ is a partial information of $F$. Then $\hat{f}(k) = {1 \over T} \sum\limits_{t=0}^{T-1} f(t) e^{-2\pi i tk / T} = {1 \over 2T} \sum\limits_{t=0}^{T-1}e^{2\pi i t(1-k) / T} + e^{-2\pi i t(1+k) / T}$ implies that 

$$\hat{f}(k) = \begin{cases}
{1 \over 2}, & \mbox{if }k =1, T-1 \\
0, & \mbox{otherwise}
\end{cases},$$ and similarly $$\hat{F}(k)= \begin{cases}
{1 \over 2}, & \mbox{if }k =2, 2T-2 \\
0, & \mbox{otherwise}
\end{cases}.$$ In the frequency domain, the length of time-series distorts the frequency information. Therefore we should consider the difference in length between $\H$ (coding process) and $\H_{\textup{new}}$ (encoding process).
\end{example}

{\color{black}The following Algorithm \ref{alg:Heuristic} is our heuristic approach to avoid the above problems.}
\begin{algorithm}[H]
\begin{algorithmic}[1]

\caption{Alternating projected gradient descent}
\label{alg:Heuristic}

\STATE \textbf{Input:} Initial point $\H_0 \in\mathbb{R}^{r \times T}$;
\vspace{0.1cm}
\STATE \textbf{Variable:} $R \in \mathbb{N}$ (the number of remaining frequencies);
\vspace{0.1cm}

\vspace{0.1cm}
\FOR{$j = 0,\cdots ,N$}

\FOR{$s = 0,\cdots ,r-1$}
\STATE
\hspace{-0.5cm} 
\begin{align}
\mathcal{K}^s_j &= \left\{k_1^s(j), \cdots, k_R^s(j), T - k_1^s(j), \cdots, T - k_R^s(j)  \right\} \label{preserving frequency indexes}\\ & \qquad \qquad \quad\text{(set of preserving frequency indexes)} \nonumber\\ 
\hspace{-0.5cm}    \H_j[s,t] &\leftarrow \sum\limits_{k \in \mathcal{K}^s_j}\widehat{\H}_j[s,k]e^{2\pi i kt / T }; \text{ delete frequencies} \text{  (projection step)} \label{frequency prior}
\end{align}
\ENDFOR
\STATE
\begin{align}
\hspace{-0.5cm}    
\H_j &\leftarrow \H_j - \gamma_j \nabla f(\H_{j}), \text{where } f(\H) = \left \lVert \begin{bmatrix} \X  \\ \sqrt{\xi}\Y \end{bmatrix} - \begin{bmatrix} \W  \\ \sqrt{\xi}{\W'} \end{bmatrix} \H \right\rVert_F^2 \text{ 
 (gradient step)}\\
\hspace{-0.5cm}    \H_j &\leftarrow \max\left\{ 0, \H_j \right\} \text{  (projection step)} \label{nonnegative prior}\\
\hspace{-0.5cm} \gamma_j &= {1 \over j+1} \cdot {\nabla f(\H_j) \over (\W^T \W +1)} 
\end{align}

\ENDFOR
\RETURN{} $\H_N$.

\end{algorithmic}
\end{algorithm}
In \eqref{frequency prior} of Algorithm \ref{alg:Heuristic}, unlike Algorithm \ref{alg:TOS}, it is not necessary to remove the same frequency at each iteration. In our experiments, for each $s \in \left\{ 0, \cdots, r-1 \right\}$, we removed a few(hyperparameter) of the frequencies with low spectrum power from the temporal data $\H[s].$ In Section \ref{sec:Experimental results}, we preserve the top $R$ frequencies of the power spectrum density for each row of $\H$ in \eqref{preserving frequency indexes}. More precisely, in \eqref{preserving frequency indexes}, we choose (positive) frequencies $\left\{k_1^s(j), \cdots,  k_R^s(j)\right\} \subseteq \left\{ 0,\cdots, \lfloor{T \over 2}\rfloor + 1 \right\}$ that satisfy $\left| \widehat{\H}_j[s,k_1^s(j)] \right| \ge \cdots \ge \left| \widehat{\H}_j[s,k_R^s(j)] \right| \ge 
 \left| \widehat{\H}_j[s,k] \right|$ for $k \in \left\{ 0,\cdots, \lfloor{T \over 2}\rfloor + 1 \right\} \setminus \left\{k_1^s(j), \cdots,  k_R^s(j)\right\}$. As mentioned in Algorithm \ref{alg:TOS}, if you want to prioritize removing frequencies, then you change the steps of \eqref{frequency prior} and \eqref{nonnegative prior} in Algorithm \ref{alg:Heuristic}.

\section{Theoretical guarantees}

\subsection{Statement of results}

In Section \ref{sec:OPT_SSNMF}, we proposed spatio-temporal matrix factorization problems, with four choices of penalization term for $\H$. In Section \ref{sec:algorithms}, we proposed two iterative algorithms for solving these problems. Namely, Algorithm \ref{alg:main} covers SSNMF with Ridge or Lasso regularization in the time domain and soft frequency regularization and   Algorithm \ref{alg:Hard2} covers SSNMF with hard frequency regularization. In this section, we state theoretical convergence guarantees for these algorithms.

\begin{theorem}[Convergence of algorithm with Ridge, Lasso, and soft frequency regularization]
\label{thm:convergence1}
With Algorithm \ref{alg:main}, $\W_j,\W'_j$ and $\H_j$ converge to stationary points  of \eqref{eq:SSNMF_main} with Ridge, Lasso and soft frequency regularization on $\H$.
\end{theorem}

The convergence results for Ridge or Lasso regularization can be proven in a similar way to soft frequency regularization. Hence we will omit the details for these cases and prove Theorem \ref{thm:convergence1} for soft frequency regularization in the following section. Next, we also establish a similar convergence result for SSNMF with hard frequency regularization using Algorithm \ref{alg:Hard2}.

\begin{theorem}[Convergence of algorithm with hard frequency regularization]\label{thm:hard_constraint_TOS}
In Algorithm \ref{alg:Hard2}, if we choose Algorithm \ref{alg:TOS} in \eqref{alg:hard sub} and \eqref{alg:hard sub_encoding}, then $\W_{j}, \W'_{j}$ and $ \H_{j}$ converge to stationary points of \eqref{eq:SSNMF_main} with hard frequency regularization. 
\end{theorem}

Algorithms \ref{alg:main} \& \ref{alg:Hard2} take the form of block coordinate descent at a high level. Because our objective function is not continuously differentiable with a convex constraint on the domain, we use the projected subgradient descent method \cite{boyd2003subgradient}.

\begin{algorithm}[H]
\begin{algorithmic}[1]
\caption{Block Coordinate Descent (BCD)}
\label{alg:BCD}

\STATE \textbf{Input:} Initial point $\boldsymbol\theta_0 = (\theta_0^1, \cdots, \theta_0^m) \in \Theta_1 \times \cdots \times \Theta_m$; $N$ (number of iterations)
\vspace{0.1cm}

\vspace{0.1cm}
\FOR{$j = 0,\cdots ,N$}
\FOR{$i = 1,\cdots ,m$}
\STATE
$$\theta^i_{j+1} = \argmin_{\theta \in \Theta_i} f(\theta^1_{j+1}, \cdots, \theta^{i-1}_{j+1}, \theta, \theta^{i+1}_{j}, \cdots, \theta^m_{j})$$
\ENDFOR
\ENDFOR
\RETURN{} $\boldsymbol\theta_N$

\end{algorithmic}
\end{algorithm}
{\color{black} Block Coordinate Descent (BCD) is an optimization technique used for nonconvex problems where the goal is to minimize a given objective function. In each iteration, BCD updates only one block of variables while keeping the other blocks constant. This implies  that each subproblem tackled by BCD is a convex problem.} But BCD does not always converge, even if $f$ is componentwise convex \cite{grippo2000convergence}. 
The challenge lies in the fact that the range of the Fourier transform is the complex domain, and $\lVert \cdot \rVert_{1,M}$ is non-differentiable.

\begin{definition}[Wirtinger derivatives \cite{da2001lectures}]
\label{not:Wirtinger derivatives}
We can regard the complex manifold $\mathbb{C}^{n} = \left\{ z_1, \cdots, z_n \right\}$ as a real manifold $\mathbb{R}^{2n} = \left\{ x_1, \cdots, x_n, y_1, \cdots, y_n \right\}$ for $z_j = x_j + i y_j$. In this context, for any point $p \in \mathbb{R}^{2n}$, the tangent space is given by $T_p \mathbb{R}^{2n} = span_{\mathbb{R}}\left\{ {\partial \over \partial x_j}\bigg|_p,  {\partial \over \partial y_j}\bigg|_p \right\}$ and the tensor product $T_p \mathbb{R}^{2n} \otimes \mathbb{C} = span_{\mathbb{C}}\left\{ {\partial \over \partial x_j}\bigg|_p,  {\partial \over \partial y_j}\bigg|_p \right\}$. To generalize complex derivatives, we introduce the Wirtinger derivatives are defined as ${\partial \over \partial z_j} = {\partial \over \partial x_j} - i {\partial \over \partial y_j}$ and ${\partial \over \partial \bar{z}_j} = {\partial \over \partial x_j} + i {\partial \over \partial y_j}$. These derivatives extend the notion of differentiation in the complex plane. It is worth noting that a $\mathbb{R}$-differentiable function $f : \mathbb{C}^{n} \rightarrow \mathbb{C}$ is $\mathbb{C}$-differentiable if and only if ${\partial f \over \partial \bar{z}} = 0$. As an example, the conjugate $\bar{z}$ does not satisfy the Cauchy-Riemann equation, which means that ${d \bar{z} \over dz}$ cannot be defined. However, we can still compute ${\partial \bar{z} \over \partial z}$, which evaluates to 0. Denote $\grad_{z} := \left( {\partial \over \partial z_1}, \cdots, {\partial \over \partial z_n} \right)$ and $\grad_{\bar{z}} := \left( {\partial \over \partial \bar{z}_1}, \cdots, {\partial \over \partial \bar{z}_n} \right)$.
\end{definition}

{\color{black}
\begin{proposition}
\label{not:real,imaginary}
${\partial \over \partial \bar{z}} \lVert z \rVert_{1,M} = {1\over 2}sign(\Re(z)) + {i \over 2}sign(\Im(z))$. Here $\Re(z)$ is the real part of $z$ and $\Im(z)$ is the imaginary part of $z$.
\end{proposition}
}

\begin{proof}
Let $z = x+iy$. ${\partial \over \partial \bar{z}} \lVert z \rVert_{1,M}  = {1 \over 2}	\left( {\partial \over \partial x} + i {\partial \over \partial y} \right)(\left| x \right| + 	\left| y \right|)= {1 \over 2}\bigl(sign(x) + i sign(y)\bigr)$. 
\end{proof}

\begin{proposition}
\label{prop:BCD sub}
$\grad_{\overline{\widehat{\H}}} \lVert \widehat{\H} \rVert_{1,M} = sign(\Re(\widehat{\H}))$.
\end{proposition}

\begin{proof}
To prove this proposition, we will consider componentwise derivatives. We have two cases of  $k=0, k = {T \over 2}$, and $1 \le k \le T-1$. This distinction is important because the $0$-th and ${T\over 2}$-th Fourier coefficients of real time-series data should be real numbers.

For each entry of $\widehat{\H}_{sk}$,
\begin{enumerate}
\item 
if $k=0$ or $k = {T \over 2}$:
\newline
${\partial \over \partial \overline{\widehat{\H}}_{sk}} \lVert \widehat{\H}_{sk} \rVert_{1,M} = {\partial \over \partial \widehat{\H}_{sk}} \lVert \widehat{\H}_{sk} \rVert_{1,M} = sign(\Re(\widehat{\H}_{sk}))$, since $\widehat{\H}_{sk} \in \R$.
\newline
    \item
    {\color{black}
    if $1 \le k \le T-1$ and $k \ne {T \over 2}$}: 
    \newline
    ${\partial \over \partial \overline{\widehat{\H}}_{sk}} \lVert \widehat{\H} \rVert_{1,M} = 
    {\color{black}
    {\partial \over \partial \overline{\widehat{\H}}_{sk}} \left( \lVert \widehat{\H}_{sk} \rVert_{1,M} + \lVert \widehat{\H}_{s,T-k} \rVert_{1,M}\right)} \\
    {\color{black}\because) \text{ The other entries of $\widehat{\H}$ are independent of $\overline{\widehat{\H}}_{sk}$.}} \\
    \overset{(a)}{=} {1\over 2}sign(\Re(\widehat{\H}_{sk})) - {i\over 2}sign(\Im(\widehat{\H}_{sk})) + {1\over 2}sign(\Re(\widehat{\H}_{s,T-k})) - {i\over 2}sign(\Im(\widehat{\H}_{s,T-k})) \\
    \overset{(b)}{=} {1\over 2} sign(\Re(\widehat{\H}_{sk})) - {i\over 2}sign(\Im(\widehat{\H}_{sk})) + {1\over 2}sign(\Re(\widehat{\H}_{sk})) + {i\over 2}sign(\Im(\widehat{\H}_{sk})) \\
    = sign(\Re(\widehat{\H}_{sk}))$,
\end{enumerate} 
where (a) follows from Proposition \ref{not:real,imaginary} and (b) uses the fact that $\Re(\widehat{\H}_{s,T-k}) = \Re(\widehat{\H}_{sk}) \text{ and } \Im(\widehat{\H}_{s,T-k}) = 
-\Im(\widehat{\H}_{sk})$.
\end{proof}

\begin{proposition}
\label{prop:subgradient}
${2 \over T} \Re(sign(\Re(\widehat{\H})) \mathcal{F}_T^{-1})$ is a subgradient of $\mathfrak{H} := \mathfrak{G} \circ (\mathfrak{F}, \overline{\mathfrak{F}})$, where $\mathfrak{F}(\H) = \widehat{\H},  \overline{\mathfrak{F}}(\H) = \overline{\widehat{\H}}$ and $\mathfrak{G}(\widehat{\H}, \overline{\widehat{\H}}) = \lVert \widehat{\H} \rVert_{1,M}$. i.e., $\mathfrak{H}(\H) = \lVert \widehat{\H} \rVert_{1,M}$. We can absorb the coefficient ${2 \over T}$ into the step size in projected subgradient descent in Lemma \ref{lemma:projected}.    
\end{proposition}

\begin{proof}    
Note that $\mathit{\Delta}\mathfrak{G}(\widehat{\H}, \overline{\widehat{\H}}) = \left\langle  \mathit{\Delta} \widehat{\H}, \grad_{\overline{\widehat{\H}}} \mathfrak{G} \right\rangle +  \left\langle  \mathit{\Delta} \overline{\widehat{\H}}, \grad_{\widehat{\H}} \mathfrak{G} \right\rangle = 2 \Re \left( \left\langle  \mathit{\Delta}\widehat{\H}, \grad_{\overline{\widehat{\H}}} \mathfrak{G} \right\rangle \right)$ \cite{li2008complex}.
\begin{align*}
& \mathit{\Delta}\mathfrak{H}(\H) = 2\Re\left(\left\langle  \mathit{\Delta}\widehat{\H}, \grad_{\overline{\widehat{\H}}} \mathfrak{G} \right\rangle \right) \\
\text{Since }\mathfrak{F} \text{ is linear,}
\\
\implies & \mathit{\Delta}\mathfrak{H}(\H) = 2\Re\left(\left\langle  \mathfrak{F} \mathit{\Delta}\H, \grad_{\overline{\widehat{\H}}}\mathfrak{G} \right\rangle \right). \\
\text{Since }\mathcal{F}_T^{*} \mathcal{F}_T = {1 \over T}\mathbf{I},& \text{ where } \mathcal{F}_T^{*} \text{ is a Hermitian matrix of } \mathcal{F}_T, \\
\implies & \mathit{\Delta}\mathfrak{H}(\H) = 2\Re\left(\left\langle   \mathit{\Delta}\H, {1 \over T}  \mathfrak{F}^{-1} \grad_{\overline{\widehat{\H}}}\mathfrak{G} \right\rangle \right). \\
\implies & \mathit{\Delta}\mathfrak{H}(\H) = \Re\left(\left\langle   \mathit{\Delta}\H, {2 \over T}(\grad_{\overline{\widehat{\H}}} \mathfrak{G}) \mathcal{F}_T^{-1}\right\rangle \right) \\
\implies & \mathit{\Delta}\mathfrak{H}(\H) = \left\langle   \mathit{\Delta}\H, {2 \over T}\Re\left((\grad_{\overline{\widehat{\H}}} \mathfrak{G}) \mathcal{F}_T^{-1} \right)\right\rangle
\end{align*} 
\end{proof}

\begin{definition}[Definition 1.4, \cite{bauschke2011convex}] 
Let $\mathcal{X}$ be a nonempty set. $f : \mathcal{X} \rightarrow [-\infty, \infty]$ is proper if $-\infty \notin f(\mathcal{X})$ and $dom f = \left\{ x \in \mathcal{X} : f(x) < \infty \right\} \ne \emptyset$.    
\end{definition}

\begin{lemma}[Convergence of BCD, Lemma 3.1, Theorem 4.1,  \cite{tseng2001convergence}]
\label{lemma:convergence}
Let $f(\theta^1,\cdots,\theta^N) = f_0(\theta^1,\cdots,\theta^N) + \sum\limits_{n=1}^N f_n(\theta^n)$. Suppose that $f_0, f_1, \cdots, f_N$ satisfy the followings
\begin{enumerate}
    \item The $dom f_0$ is open and $f_0$ has a directional derivative in all direction on $dom f_0$.
    \item The set $X^0 = 	\left\{(\theta^1,\cdots,\theta^N): f(\theta^1,\cdots,\theta^N) \le f (\theta^1_0,\cdots,\theta^N_0)\right\}$ is
compact, and $f$ is continuous on $X^0$.
    \item The mapping $\theta^n \mapsto f(\theta^1,\cdots,\theta^N)$ has at most one minimum for $n \ge 2$.
\end{enumerate}
Then, for $\left\{(\theta_k^1, \cdots, \theta_k^N) \right\}_{k \in \mathbb{N}}$, every cluster point is a stationary point of $f$ . 
\end{lemma}

\begin{lemma}[Projected subgradient method, \cite{boyd2003subgradient}]
\label{lemma:projected}
Consider the following constraint convex optimization problem
\begin{equation}
\label{Projected subgradient method}
\argmin_{x \in \mathcal{C}} f(x)    
\end{equation}
where $f:\R^n \rightarrow \R$ is convex and $\mathcal{C} \subseteq \R^n$ is convex.    
The projected subgradient method is given by $x_{k+1} = P(x_{k} - \alpha_k g_{(k)})$, where $P$ is the projection on $\mathcal{C}$ and $\alpha_k$ is the $k$th step size and $g_{(k)}$ is a subgradient of $f$ at $x_{k}$. Then $x_{k}$ converges to the solution of \eqref{Projected subgradient method}.
\end{lemma}

Note that Problem \eqref{eq:SSNMF_main} is a nonconvex problem. Therefore, finding the global minimum of this problem is extremely difficult. Alternatively, to tackle this issue, we aim to find the stationary points. The Subproblems \eqref{eq:coding_H}, \eqref{eq:coding_W0} and \eqref{eq:coding_W1} are all convex optimization problems. We will particularly focus on examining \eqref{eq:coding_H}. Since $\H \mapsto \lVert \widehat{\H} \rVert_{1,M}$ is not differentiable, we use projected `sub'gradient descent algorithm to solve \eqref{eq:coding_H}. Proposition \ref{prop:subgradient} will be used to solve it.

\begin{proof}{of Theorem \ref{thm:convergence1}. }
\newline
(BCD convergence) 
In Lemma \ref{lemma:convergence}, set $\theta^1 = \H, \theta^2 = \W, \theta^3 = \W'$ and $f_0(\H,\W,$ 
$\W') = \lVert \X - \W\H \rVert_F^2 + \xi \lVert \Y - \W'\H \rVert_F^2, f_1(\H) = \lambda \lVert \H \mathcal{F}_T \rVert_{1,M} + \iota_{\mathcal{G}}(\H), f_2(\W) = \lambda_1 \lVert \W \rVert_F^2$ and $f_3(\W') = \lambda_2 \lVert \W' \rVert_F^2$ , where $\xi, \lambda, \lambda_1, \lambda_2 >0$, $ \mathcal{G}= \bigl\{ \H \in \mathbb{R}^{r\times T} : \H \ge 0 \bigr\}$ and $\iota_{\mathcal{G}}$ is an indicator function. Here, we introduce $L2$-regularization terms for $\W$ and $\W'$ to satisfy conditions 2 and 3 in Lemma 1. Note that $\lambda_1$ and $\lambda_2$ can be selected as arbitrarily small positive numbers; thus, in Algorithm 1, we can disregard the $L2$-regularization term.

Now, we check the conditions in the Lemma \ref{lemma:convergence}.

\begin{enumerate}
    \item Since $dom f_0 = \R^{r\times T} \times \R^{d \times r} \times \R^{d \times r}$ is whole space, it is open. Clearly, $f_0$ is continuously differentiable, it has a directional derivative in all directions.
    \item Suppose we choose an initial point $\H_0 \in \mathcal{G}$. Then $M:= f(\H_0, \W_0, \W'_0)$ has a finite value. The set $X^0 = 	\bigl\{(\H,\W,\W'): f(\H,\W,\W') \le M \bigr\} = f^{-1}([0,M]) (\because f \ge 0)$. Note that $f$ on $X^0$ is $\lVert \X - \W\H \rVert_F^2 + \xi \lVert \Y - \W'\H \rVert_F^2 + \lambda \lVert \H \mathcal{F}_T \rVert_{1,M} +  \lambda_1 \lVert \W \rVert_F^2 +  \lambda_2 \lVert \W' \rVert_F^2$. Since $\lVert \H \mathcal{F}_T \rVert_{1,M}, \lVert \W \rVert_F$ and $\lVert \W' \rVert_F$ are bounded, the set $X^0$ becomes bounded. Clearly, $f$ is continuous on $X^0$, this implies that $X^0 = f^{-1}([0,M])$ is closed; therefore, it is compact.

    \item Note that if the Hessian of $f$ is positive definite, then $f$ is strictly convex \cite{grasmair2016basic}. We can calculate the following (see \cite{petersen2008matrix})
\begin{equation}
{\partial f  \over \partial \W} = 2(\W \H \H^T - \X \H^T)+ 2\lambda_1 \W, {\partial^2 f  \over \partial \W^2} = 2\H \H^T + 2\lambda_1 \mathbf{I},   
\end{equation}
\begin{equation}
{\partial f  \over \partial \W'} = 2\xi(\W' \H \H^T - \X \H^T)+ 2\lambda_2 \W', {\partial^2 f  \over \partial \W'^2} = 2\xi\H \H^T + 2\lambda_2 \mathbf{I},    
\end{equation}
where $\mathbf{I}$ is an identity matrix. Since $\lambda_1, \lambda_2 >0$, ${\partial^2 f  \over \partial \W^2}$ and ${\partial^2 f  \over \partial \W'^2}$ are positive definite. Therefore, $\W \mapsto f(\H, \W, \W')$ and $\W' \mapsto f(\H, \W, \W')$ are strictly convex and have at most one minimum.
\end{enumerate}
(Solving subproblem via BCD): The Fourier transform $\H\mapsto \widehat{\H}$ is linear and $\lVert \cdot \rVert_{1,M}$ is a convex function. Consequently, the mapping $\H\mapsto \lVert \widehat{\H} \rVert_{1,M} = \lVert \H\mathcal{F}_T \rVert_{1,M}$ is also convex and it satisfies all the hypotheses of Lemma \ref{lemma:projected}. We have computed its subgradient in Proposition \ref{prop:subgradient}. Therefore, we can solve the subproblem using the projected subgradient descent: 
\begin{equation}
\argmin\limits_{\H \in \R_{\ge 0}^{r\times T}} \lVert \X - \W\H \rVert_F^2 + \xi \lVert \Y - \W'\H \rVert_F^2 + \lambda \lVert \H \mathcal{F}_T \rVert_{1,M}.
\end{equation}
The other subproblems: 
\begin{equation}
\argmin\limits_{\W \in \R^{d\times T}} \lVert \X - \W\H \rVert_F^2 + \xi \lVert \Y - \W'\H \rVert_F^2 + \lambda \lVert \H \mathcal{F}_T \rVert_{1,M}
\end{equation} and 
\begin{equation}
\argmin\limits_{\W' \in \R^{d\times T}} \lVert \X - \W\H \rVert_F^2 + \xi \lVert \Y - \W'\H \rVert_F^2 + \lambda \lVert \H \mathcal{F}_T \rVert_{1,M}
\end{equation}
are usual normal equation and its solution is well-known.
\newline
\newline
Therefore  $\W_j$, $\W'_j$ and $\H_j$ converge to stationary points in Algorithm \ref{alg:main}.
\end{proof}

\begin{remark}
\textit{To solve the subproblem \eqref{eq:coding_H}, we apply the following gradient-projection steps.}
\textit{\begin{align}
\label{eq:gradient step}
\text{ (Gradient step)} \quad \H_{j} &\leftarrow \H_j - \alpha_{j} \nabla f(\H_j), \text{where}\\
\hspace{-0.5cm} f(\H_j) &=
({\W_{j-1}^T}\W_{j-1}\H_{j-1} - {\W_{j-1}^T}\X) \nonumber \\
\hspace{-0.5cm} &+ \xi ({{\W'}_{j-1}^T}{\W'}_{j-1}\H_{j-1} - {{\W'}_{j-1}^T}\Y) \nonumber \\ 
\hspace{-0.5cm} &+ \lambda \Re(sign(\Re(\widehat{\H})) \mathcal{F}_T^{-1})&  \\
\label{eq:projection step}
\text{ (Projection step)} \quad  \H_{j} & \leftarrow  \max\left\{ 0, \H_j - \alpha_{j} \nabla f(\H_j) \right\}.  
\end{align}}

\textit{This can be interpreted from a frequency domain perspective.
\begin{align}
\widehat{\H}_j &\leftarrow \widehat{\H}_j - \alpha_{j} \nabla f(\widehat{\H}_j), \text{ where} \label{eq:updateremark}\\
\hspace{-0.5cm}  f(\widehat{\H}_j) &=
({\W_{j-1}^T}\W_{j-1}{\widehat{\H}}_{j-1} - {\W_{j-1}^T}\widehat{\X}) \nonumber \\
\hspace{-0.5cm} &+ \xi ({{\W'}_{j-1}^T}{\W'}_{j-1}{\widehat{\H}}_{j-1} - {{\W'}_{j-1}^T}\widehat{\Y}) \nonumber \\ 
\hspace{-0.5cm} &+ \lambda \Re\left(sign(\Re(\widehat{\H}))\right), \label{eq:signremark}
\end{align}
where Proposition \ref{prop:BCD sub} is used in \eqref{eq:signremark}. By applying the inverse Fourier transform to update rule \eqref{eq:updateremark}, we obtain the update rule in the time domain.}
\end{remark}

Now, consider Theorem \ref{thm:hard_constraint_TOS}.

\begin{proposition}[Corollary 1 and Theorem 4, \cite{yurtsever2021three}]
\label{prop:three operator}
If $f : \mathbb{R}^{r\times T} \rightarrow \mathbb{R}$ and $g = \iota_{\mathcal{G}},h_R = \iota_{\mathcal{H}} : \mathbb{R}^{r \times T} \rightarrow \mathbb{R} \cup 	\left\{ \infty \right\}$ are proper, lower semi-continuous and convex, where $\mathcal{G}$ and $\mathcal{H} \subseteq \mathbb{R}^{n}$ are convex and $\iota_{\mathcal{G}}, \iota_{\mathcal{H}}$ are indicator functions. Let $\overline{\H}_N$ be output of Algorithm \ref{alg:TOS}, then 
\begin{align}
f(\overline{\H}_N) - \min f \commHL{=} \tilde{\mathcal{O}} \left( {1 \over N} \right), \\ 
\textup{dist}(\overline{\H}_N , \mathcal{H}) \commHL{=} \tilde{\mathcal{O}} \left( {1 \over N} \right).
\end{align}

\end{proposition}

We will show that the matrix factorization with two constraints, i.e. 1) the nonnegativity constraint of $\H$ and 2) the constraint that $\H$ has no specific frequencies and also satisfies the hypothesis of Proposition \ref{prop:three operator}.

\begin{lemma}[Example 1.25, \cite{bauschke2011convex}]
\label{lemma:indicator function} Let $\mathcal{X}$ be a Hausdorff space. The indicator function of a set $C \subset \mathcal{X}$, i.e. $\iota_{C} : \mathcal{X} \rightarrow [-\infty, \infty] : x \mapsto \begin{cases}
0, & \mbox{if }x \in C \\
\infty, & \mbox{Otherwise}
\end{cases}$ is lower semi-continuous if and only if $C$ is closed.
\end{lemma}

\begin{proof}{of Theorem \ref{thm:hard_constraint_TOS}.} This proof is similar to the proof of Theorem \ref{thm:convergence1}.
\newline
(BCD convergence) 
Let $\mathcal{G} = \bigl\{ \H \in \mathbb{R}^{r\times T} : \H \ge 0 \bigr\}$ and $\mathcal{H} = \mathcal{F}_{T}^{-1} (\mathbb{C}^{r \times R} \times \left\{ 0 \right\}^{T-R})$. In Lemma \ref{lemma:convergence}, set $\theta^1 = \H, \theta^2 = \W, \theta^3 = \W'$ and $f_0(\H,\W,$ 
$\W') = \lVert \X - \W\H \rVert_F^2 + \xi \lVert \Y - \W'\H \rVert_F^2, f_1(\H) = \lambda \lVert \H \rVert_F^2 + \iota_{\mathcal{G} \bigcap \mathcal{H}}(\H), f_2(\W) = \lambda_1 \lVert \W \rVert_F^2$ and $f_3(\W') = \lambda_2 \lVert \W' \rVert_F^2$, where $\xi, \lambda, \lambda_1, \lambda_2 >0$. If we choose an initial point $\H_0 \in \mathcal{G} \bigcap \mathcal{H}$, then $f$ on $X^0$ is $\lVert \X - \W\H \rVert_F^2 + \xi \lVert \Y - \W'\H \rVert_F^2 + \lambda \lVert \H \rVert_{F}^2 +  \lambda_1 \lVert \W \rVert_F^2 +  \lambda_2 \lVert \W' \rVert_F^2$. The remainder of the proof is similar to that of Theorem \ref{thm:convergence1}.
\newline
\newline
(How to solve subproblem in BCD)
Set $f,g$ and $h_R$ defined in \eqref{eq:f} - \eqref{eq:h}. Since $\mathcal{G}$ and $\mathcal{H}$ are closed convex sets in $\mathbb{R}^{r \times T}$, by Lemma \ref{lemma:indicator function}, $g$ and $h_R$ are lower semi-continuous. Since the indicator function on convex set is convex map, $g$ and $h_R$ are convex maps. Clearly, $g$ and $h_R$ are proper maps. Therefore $f,g$ and $h_R$ satisfy the all hypothesis of Proposition \ref{prop:three operator}.
Therefore  $\W_j$, $\W'_j$ and $\H_j$ converge to stationary points in Algorithm \ref{alg:Hard2}.
\end{proof}

\section{Analysis on synthetic data}

The following proposition states that the presence of a mismatch term between $\widehat{\X}$ and $\widehat{\Y}$ can disrupt the inference of the periodic pattern of $\X$. Here $S$ is the number of auxiliary data.

\begin{proposition}\label{prop:synthetic_code_closed_form}
\label{Fourier difference}
Let $\overline{\X} = \begin{bmatrix} \X  \\ \sqrt{\xi}\Y \end{bmatrix}$ and $\overline{\W} = \begin{bmatrix} \W  \\ \sqrt{\xi}\W' \end{bmatrix}$.
If we let $\X_{mn} = \sum\limits_{l=0}^{T-1} \widehat{\X}_{ml} e^{2\pi iln\over T}$ and $\Y_{mn} = \sum\limits_{l=0}^{T-1} \widehat{\Y}_{ml} e^{2\pi iln\over T}$, then the solution of $\min\limits_{\H} \lVert \overline{\X} - \overline{\W}\H \rVert_{F}^2$ is given by the following.
\begin{align}
\H_{kj} &= \sum\limits_{s=0}^{d-1} \left[\left((\overline{\W}^T \overline{\W})^{-1} \overline{\W}^T\right)_{ks}  + \sum\limits_{p=0}^{S-1} \sqrt{\xi}   \left((\overline{\W}^T \overline{\W})^{-1} \overline{\W}^T\right)_{k,s+pd}\right] \X_{sj}  \nonumber \\
&+ \sum\limits_{s=0}^{d-1} \sum\limits_{p=0}^{S-1} \sum\limits_{l=0}^{T-1} \left((\overline{\W}^T \overline{\W})^{-1} \overline{\W}^T\right)_{k,s+pd} \sqrt{\xi} \left((\widehat{\Y_p})_{sl} -\widehat{\X}_{sl}\right)e^{2\pi ilj\over T}.
\end{align}
\end{proposition}

\begin{proof}

The solution to $\min\limits_{\H} \lVert \overline{\X} - \overline{\W}\H \rVert_{F}^2$ is well-known as the normal equation, and its solution is $\H = (\overline{\W}^T \overline{\W})^{-1} \overline{\W}^T \overline{\X}$.
\begin{align*}
\H_{kj} &= \sum\limits_{s=0}^{d+dS-1} \left((\overline{\W}^T \overline{\W})^{-1} \overline{\W}^T\right)_{ks} \overline{\X}_{sj} \\
&= \sum\limits_{s=0}^{d-1} \left((\overline{\W}^T \overline{\W})^{-1} \overline{\W}^T\right)_{ks}\X_{sj} + \sum\limits_{s=0}^{dS-1} \left((\overline{\W}^T \overline{\W})^{-1} \overline{\W}^T\right)_{k,s+d} \sqrt{\xi} \Y_{sj} \\
&= \sum\limits_{s=0}^{d-1} \left((\overline{\W}^T \overline{\W})^{-1} \overline{\W}^T\right)_{ks}\X_{sj} + \sum\limits_{s=0}^{d-1} \sum\limits_{p=0}^{S-1} \left((\overline{\W}^T \overline{\W})^{-1} \overline{\W}^T\right)_{k,s+pd} \sqrt{\xi} (\Y_p)_{sj}. \\
\end{align*}
Using the relations $\X_{sj} = \sum\limits_{l=0}^{T-1} \widehat{\X}_{sl} e^{2\pi ilj\over T}$ and $({\Y_p})_{sj} = \sum\limits_{l=0}^{T-1} (\widehat{\Y_p})_{sl} e^{2\pi ilj\over T}$, we have
\begin{align*}
\H_{kj} &= \sum\limits_{s=0}^{d-1} \sum\limits_{l=0}^{T-1}  \left((\overline{\W}^T \overline{\W})^{-1} \overline{\W}^T\right)_{ks} \widehat{\X}_{sl} e^{2\pi ilj\over T} \\ 
&+ \sum\limits_{s=0}^{d-1}
\sum\limits_{p=0}^{S-1}
\sum\limits_{l=0}^{T-1}
\left((\overline{\W}^T \overline{\W})^{-1} \overline{\W}^T\right)_{k,s+pd} \sqrt{\xi} (\widehat{\Y_p})_{sl} e^{2\pi ilj\over T}.
\end{align*}
To see the disparity term between $\widehat{\X}$ and $\widehat{\Y}$, we rearrange the above as the following, 
\begin{align*}
\H_{kj} &= \sum\limits_{s=0}^{d-1} \sum\limits_{l=0}^{T-1} \left((\overline{\W}^T \overline{\W})^{-1} \overline{\W}^T\right)_{ks} \widehat{\X}_{sl} e^{2\pi ilj\over T} \\
&+ \sum\limits_{s=0}^{d-1}  \sum\limits_{p=0}^{S-1}\sum\limits_{l=0}^{T-1} \left((\overline{\W}^T \overline{\W})^{-1} \overline{\W}^T\right)_{k,s+pd} \sqrt{\xi} \widehat{\X}_{sl} e^{2\pi ilj\over T} \\
&+ \sum\limits_{s=0}^{d-1}  \sum\limits_{p=0}^{S-1}\sum\limits_{l=0}^{T-1}\left((\overline{\W}^T \overline{\W})^{-1} \overline{\W}^T\right)_{k,s+pd} \sqrt{\xi} \left((\widehat{\Y_p})_{sl} -\widehat{\X}_{sl}\right)e^{2\pi ilj\over T}  \\
&= \sum\limits_{s=0}^{d-1} \left[\left((\overline{\W}^T \overline{\W})^{-1} \overline{\W}^T\right)_{ks}  + \sum\limits_{p=0}^{S-1} \sqrt{\xi}   \left((\overline{\W}^T \overline{\W})^{-1} \overline{\W}^T\right)_{k,s+pd}\right] \\
&\times \left(\sum\limits_{l=0}^{T-1} \widehat{\X}_{sl} e^{2\pi ilj\over T} \right)  \\
&+ \sum\limits_{s=0}^{d-1} \sum\limits_{p=0}^{S-1} \sum\limits_{l=0}^{T-1} \left((\overline{\W}^T \overline{\W})^{-1} \overline{\W}^T\right)_{k,s+pd} \sqrt{\xi} \left((\widehat{\Y_p})_{sl} -\widehat{\X}_{sl}\right)e^{2\pi ilj\over T}.
\end{align*}
Now the inverse Fourier transform yields
\begin{align*}
\H_{kj}&= \sum\limits_{s=0}^{d-1} \left[\left((\overline{\W}^T \overline{\W})^{-1} \overline{\W}^T\right)_{ks}  + \sum\limits_{p=0}^{S-1} \sqrt{\xi}   \left((\overline{\W}^T \overline{\W})^{-1} \overline{\W}^T\right)_{k,s+pd}\right] \X_{sj}  \\
&+ \sum\limits_{s=0}^{d-1} \sum\limits_{p=0}^{S-1} \sum\limits_{l=0}^{T-1} \left((\overline{\W}^T \overline{\W})^{-1} \overline{\W}^T\right)_{k,s+pd} \sqrt{\xi} \left((\widehat{\Y_p})_{sl} -\widehat{\X}_{sl}\right)e^{2\pi ilj\over T}.
\end{align*}
\end{proof}

\begin{figure}[!]
    \centering
    \includegraphics[width=1 \linewidth]{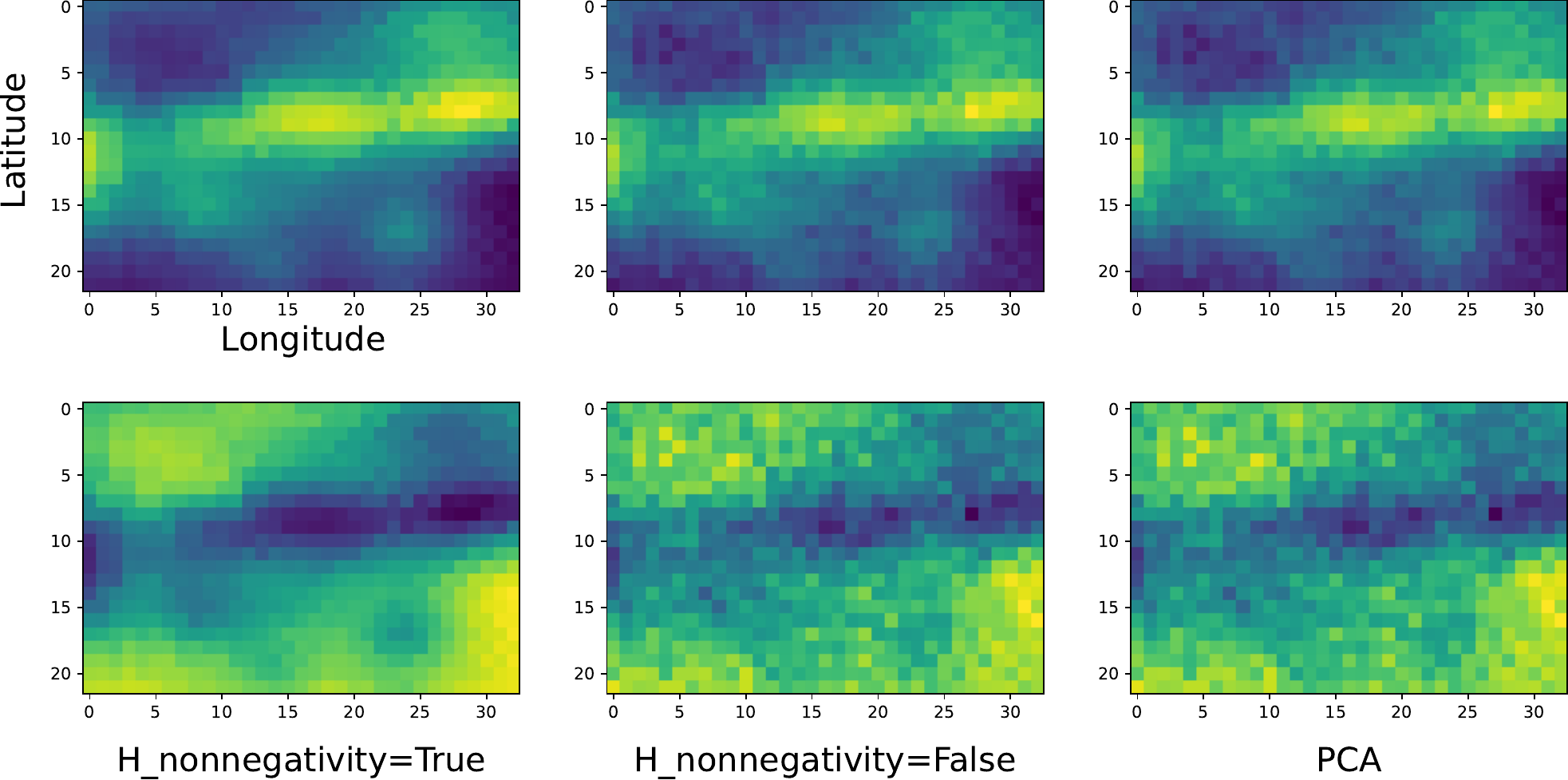}
    \newline
    \newline
    \includegraphics[width=1 \linewidth]{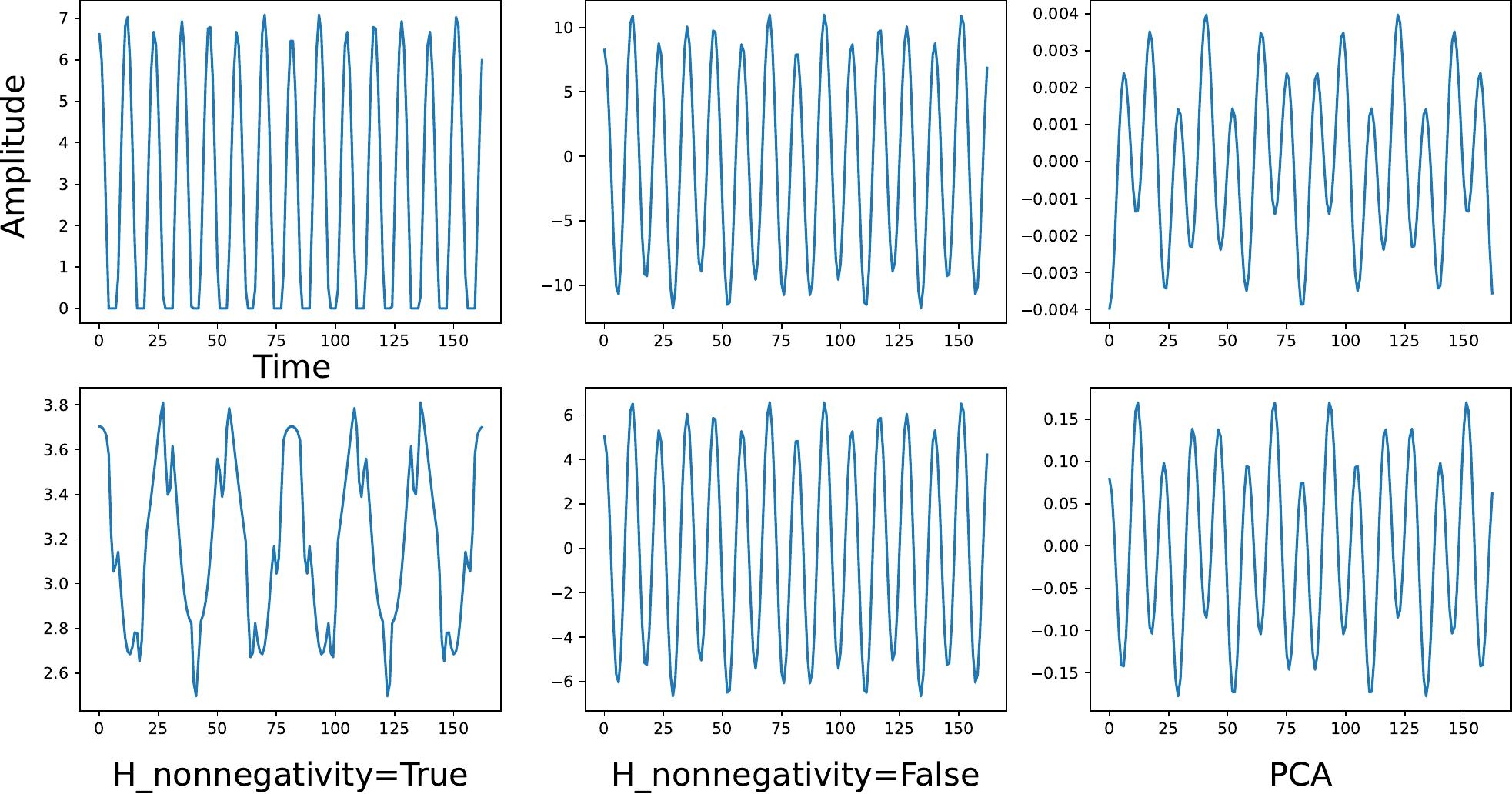}
    \newline
    \newline
    \includegraphics[width=1 \linewidth]{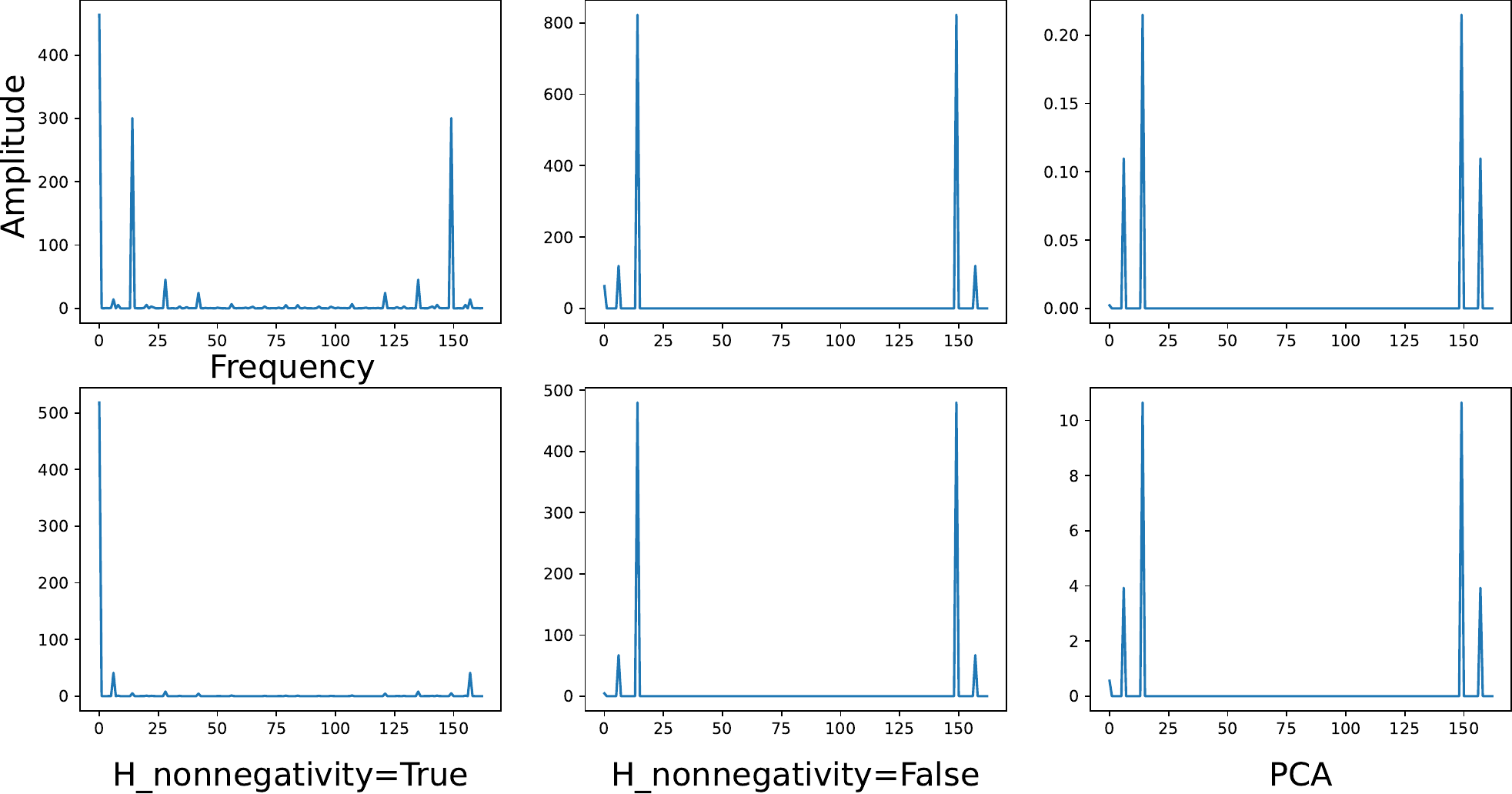}
    \caption{Top: spatial patterns of $\X$ in Example \ref{ex1}.
    First(second) row shows the first(second) column of $\W$. Middle: temporal patterns of $\H$. First(second) row is the first(second) row of $\H$. Bottom: spectrums of temporal patterns of $\H$. First(second) row is the Fourier transform of the first(second) row of $\H$. The left column shows the results when the nonnegativity constraint is enforced to
$\H$, the middle column without the nonnegativity constraint and the right
column with PCA.}
    \label{fig:synthetic experiment}
\end{figure}

\begin{example}\label{ex1}
Set $\X_{ij} = \X_{i0}$ $ \cdot \left[\cos\left( {2\pi \cdot 14 j \over 163} \right) + \cos\left( {2\pi \cdot 6 j \over 163} \right)\right] + \epsilon(0,1), (\Y_{0})_{ij} = (\Y_0)_{i0} $ $ \times \cos\left( {2\pi \cdot 14 j \over 163} \right) + \epsilon(0,\sigma)$ and $(\mathbf{Y}_1)_{ij} = (\mathbf{Y}_1)_{i0} \cos\left( {2\pi \cdot 6 j \over 163} \right) + \epsilon(0,\sigma)$ where $\sigma \in \left\{ 0,1,2,3,4,5 \right\}$ and $\epsilon(m,\sigma)$ is a Gaussian noise with the mean $m$ and the standard deviation $\sigma$. We apply the matrix factorization based on the following three cases \lowercase\expandafter{\romannumeral1}) with the constraint on $\H$, \lowercase\expandafter{\romannumeral2}) without any constraint on $\H$, \lowercase\expandafter{\romannumeral3}) with the Principal Component Analysis(PCA).  Figure \ref{fig:synthetic experiment} shows the spatial (top rows) and temporal (middle rows) patterns and the corresponding spectrums (bottom rows). The left, middle, and right columns correspond to the matrix factorization method with the nonnegativity constraint enforced, without the nonnegativity constraint enforced, and with the PCA approach, respectively. {\color{black} In this example, we use synthetic data for $\X$ that has two temporal patterns with frequencies of $6$ and $14$. As shown in  Fig. \ref{fig:synthetic experiment}, when the nonnegativity constraint is enforced to $\H$, we obtain clear spatial patterns, and each row of $\H$ captures different temporal patterns. Here note that the result may depend on the initial values assigned to $\H$ in BCD algorithm}.  On the other hand, when we do not impose the nonnegative constraint on $\H$ (or when using PCA), their temporal patterns cannot separate those two temporal frequency patterns, 6 and 14. In the time domain, we can observe that $\H$ has nonnegative values in the first column.
\end{example}

{\color{black}\begin{example}
[Comparison of Frobenius norm, L1-norm, and Minkowski 1-norm using projected (sub)gradient descent] 
\label{ex:three norms}
In this example, we compare the behaviors of $\H$ in both the time  and frequency domains when applying projected (sub)gradient descent method with the Frobenius norm, L1-norm, and Minkowski 1-norm of $\H$. Consider a time-series data $\H_0[k]$ $= \cos(10\pi \cdot {k \over 30}) + \epsilon_k$, where $\epsilon_k \in [0,1]$ is uniform noise for $k=0,1,\cdots,29$.
In order to minimize the Frobenius, L1-, and Minkowski 1-norms$, i.e.  
\lVert \H \rVert_{F}^2$,
$\lVert \H \rVert_{1}$, and 
$\lVert \H \rVert_{1,M}$, respectively, 
we apply the update rules with the projected subgradient descent method to each norm as follows: 

\textbf{Frobenius norm}: 
$\H_{j+{1\over 2}} = \H_j - \alpha_{j} \H_j$ and $\H_{j+1} = \max\bigl\{ 0, $ $\H_{j + {1\over 2}} \bigr\}$.

\textbf{L1-norm}: 
$\H_{j+{1\over 2}} = \H_j - \beta_{j} (sign(\H_j))$ and $\H_{j+1} = \max\bigl\{ 0, $ $\H_{j + {1\over 2}} \bigr\}$.

\textbf{Minkowski 1-norm}: 
$\H_{j+{1\over 2}} = \H_j - \gamma_{j} (sign(\Re(\hat{\H}_j))\mathcal{F}_T^{-1})$ and $\H_{j+1} = \max\left\{ 0, \H_{j + {1\over 2}} \right\}$.

In this experiment, we set $\alpha=0.1, \beta=0.1$ and $\gamma=0.5$. For a fair comparison, we iterate the projected (sub)gradient method until $\lVert \H \rVert_F < 3$. Figure \ref{fig:synthetic} displays $\H$ after projected (sub)gradient descent, with the Frobenius norm (blue), the L1-norm (green), and the Minkowski 1-norm (red). The left figure in Fig. \ref{fig:synthetic} shows the data in the time domain and the right figure in the frequency domain. In this example, the solution is simply $\H=0$. We want to observe how $\H_j$ approaches $0$ during the gradient descent process. Note that the iteration goes  until $\lVert \H \rVert_F < 3$ and according to Parseval's identity, $\lVert \widehat{\H} \rVert_F$ are same for three cases. The difference among  these three cases in the time and frequency domains is the amplitude distribution.
In the case of the L1-norm, as expected, it exhibits significant  sparsity in the time domain. Although the Minkowski 1-norm does not achieve the desired sparsity in the frequency domain, it shows a tendency to yield smaller amplitudes of frequencies compared to other norms. In this experiment, we expect the 5th frequency to be dominant, while the other frequencies are considered noise originating from uniform noise. The Minkowski 1-norm shows an effect of increasing the amplitude of the 0th frequency while reducing the amplitudes of undominated (noise) frequencies. This effect remains consistent even when experimenting with different random seeds for uniform noise. The experimental results show that the Minkowski 1-norm reduces the amplitude of the undominated frequencies more effectively than the Frobenius and L1-norms. 
That is, the soft frequency regularization  achieves our goal partially as stated in Section \ref{sec:Hard constraints}.

\begin{figure}[!]
    \centering
    \includegraphics[width=0.45 \linewidth]{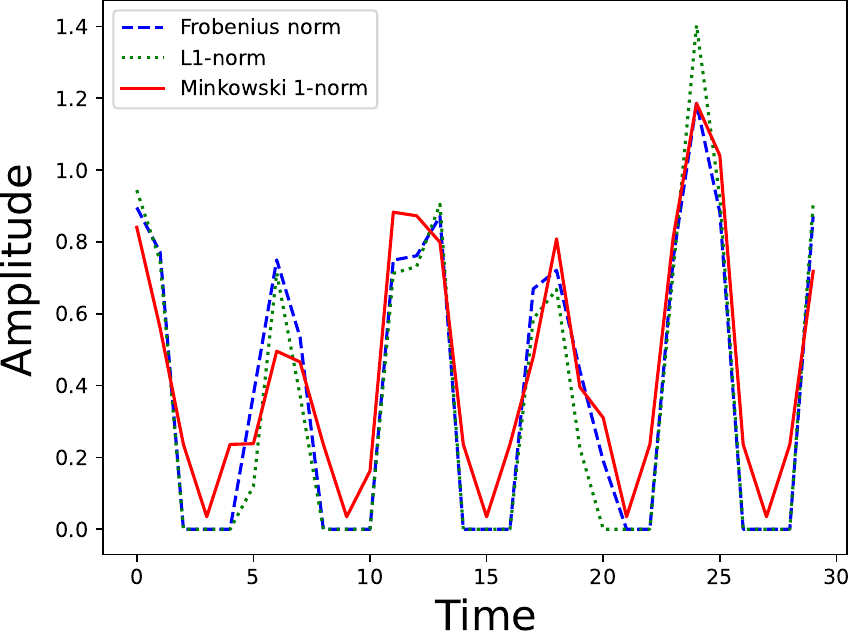}
    \includegraphics[width=0.45 \linewidth]{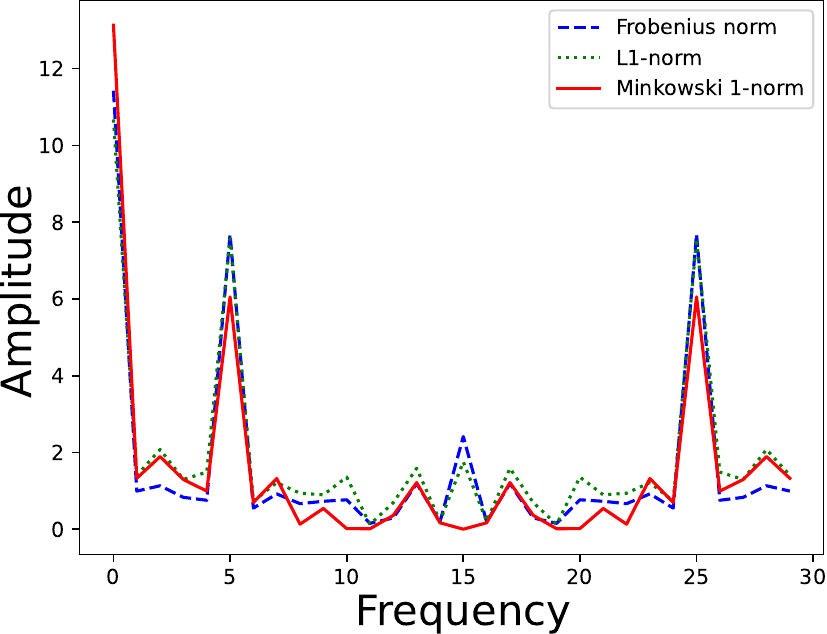}
    \caption{Behaviors under the projected (sub)gradient descent in Example \ref{ex:three norms}. $\H$ with the Frobenius norm (blue), the L1-norm (green), and the Minkowski 1-norm (red). Left: $\H$ in the time domain. Right: $\left| \widehat{\H} \right|$ in the frequency
domain.}
    \label{fig:synthetic}
\end{figure}

\end{example}}

\section{Experimental results}
\label{sec:Experimental results}

\subsection{GRACE data}\label{sec:GRACE}
Gravity Recovery and Climate Experiment (GRACE) satellites measure the change of the gravitational field on Earth. From this measurement, we can obtain the total water storage anomaly. Our dataset $\mathcal{X}$ represented as a 3-tensor, and each entry value $\mathcal{X}[m,n,t]$ is the variation in total water storage compared to the temporal average during Jan. 2004 - Dec. 2009 at a latitude of $m$, longitude of $n$, and time (months) of $t$\footnote{\url{https://edo.jrc.ec.europa.eu/documents/factsheets/factsheet_grace_tws_anomaly.pdf}}. The measurement unit is centimeters (cm). By considering variations in total water storage, we can remove static mass and infer the redistribution of water. Matrix factorization is one method used to extract latent patterns from the data. In our research, matrix factorization is used to separate spatio-temporal data into spatial latent figures and temporal latent figures. This approach enables us to analyze temporal patterns. The nonnegative constraint helps us to obtain more interpretable spatial atoms. To analyze the temporal patterns, we consider a soft constraint in the frequency domain. For the reconstruction of the total water storage anomaly (TWSA)\footnote{CSR GRACE/GRACE-FO RL06 Mascon Solutions, \url{https://www2.csr.utexas.edu/grace/RL06_mascons.html} \cite{save2016high, save2020csr}}, we use auxiliary data that was also utilized in \cite{sun2020reconstruction}: precipitation\footnote{ERA5 monthly averaged data on single levels, \url{https://cds.climate.copernicus.eu/cdsapp\#!/dataset/reanalysis-era5-single-levels-monthly-means?tab=overview}\label{note1} \cite{hersbach2019era5}}, temperature\textsuperscript{\ref{note1}}, and Noah TWS\footnote{GLDAS Noah Land Surface Model, \url{https://disc.gsfc.nasa.gov/datasets/GLDAS_NOAH025_M_2.1/summary?keywords=gldas} \cite{beaudoing25nasa, rodell2004global}}. Noah TWS refers to the traditional method of restoring TWSA before the GRACE era. We divide the dataset into the training set (132 months, Apr. 2002 - Jan. 2014) and the test set (31 months, Feb. 2014 - Jun. 2017). The train set consists of the data for the 142 months, and the test set for the remaining 41 months. However, due to technical issues, there are missing data in both sets, occurring intermittently every 10 months (06/2002, 07/2002, 06/2003, 01/2011, 06/2011, 05/2012, 10/2012, 03/2013, 08/2013, 09/2013, 02/2014, 07/2014, 12/2014, 06/2015, 10/2015, 11/2015, 04/2016, 09/2016, 10/2016, 02/2017). These missing data have been excluded as described in \cite{sun2020reconstruction}. We chose 17 river basin data to assess our proposed method, that is, the areas of Amazon, Congo, Ganges, Indus, Huanghe, Volga, Parana, Lena, Mackenzie, Ob, Yenisei, Nile, Yangtze, Indigirka, Zambezi, Mississippi, Murray\footnote{
The latitude and longitude information for the river basin was obtained from \url{https://hydro.iis.u-tokyo.ac.jp/~taikan/TRIPDATA/Data/rivers.idx} \cite{oki1998design}.}.

We focus on the temporal patterns present in spatio-temporal data. In our experiments with the GRACE data, we expect to find a spatial pattern that exhibits an annual cycle with a period of 12 months. In this study, we utilize the discrete Fourier transform on the time-series data obtained through SSNMF to uncover the temporal patterns in GRACE data. By preserving specific frequencies selectively in the time-series corresponding to each spatial basis, we can enhance the interpretability of the periodic patterns. To achieve this, we use the method proposed in the pervious section, i.e. the soft/hard constraints in the frequency domain and compare the results with the results by the Lasso/Ridge regression methods. Furthermore, we compare the methodology with the existing methods such as Deep Neural Network(DNN), Multiple Linear Regression(MLR), Seasonal ARIMA with eXogenous variables(SARIMAX) \cite{cools2009investigating}, and assess its forecasting performance, demonstrating comparable results.

\begin{figure}[!]
    \centering
    \includegraphics[width=1 \linewidth]{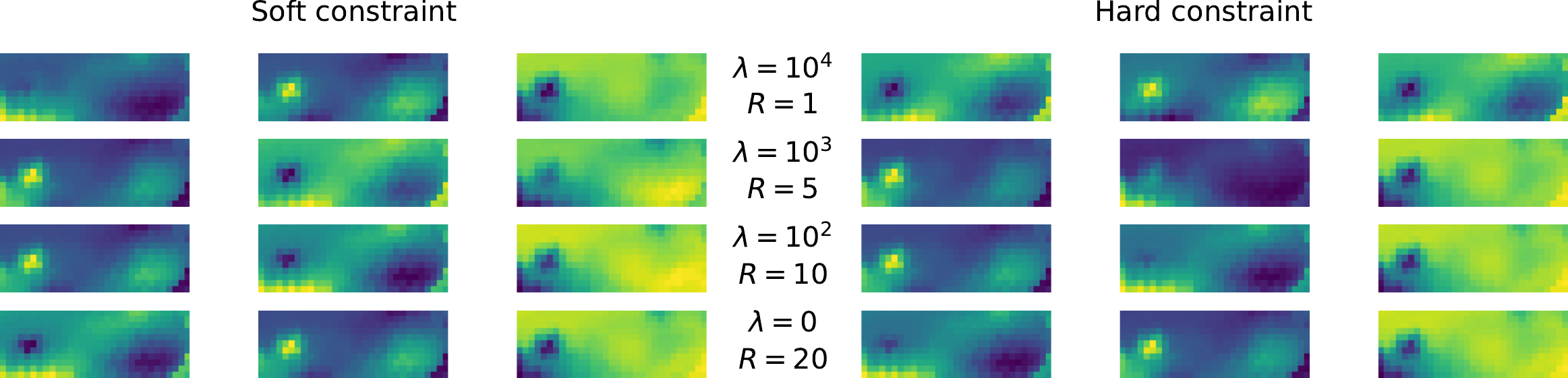}
    \newline
    \newline
    \newline
\includegraphics[width=1 \linewidth]{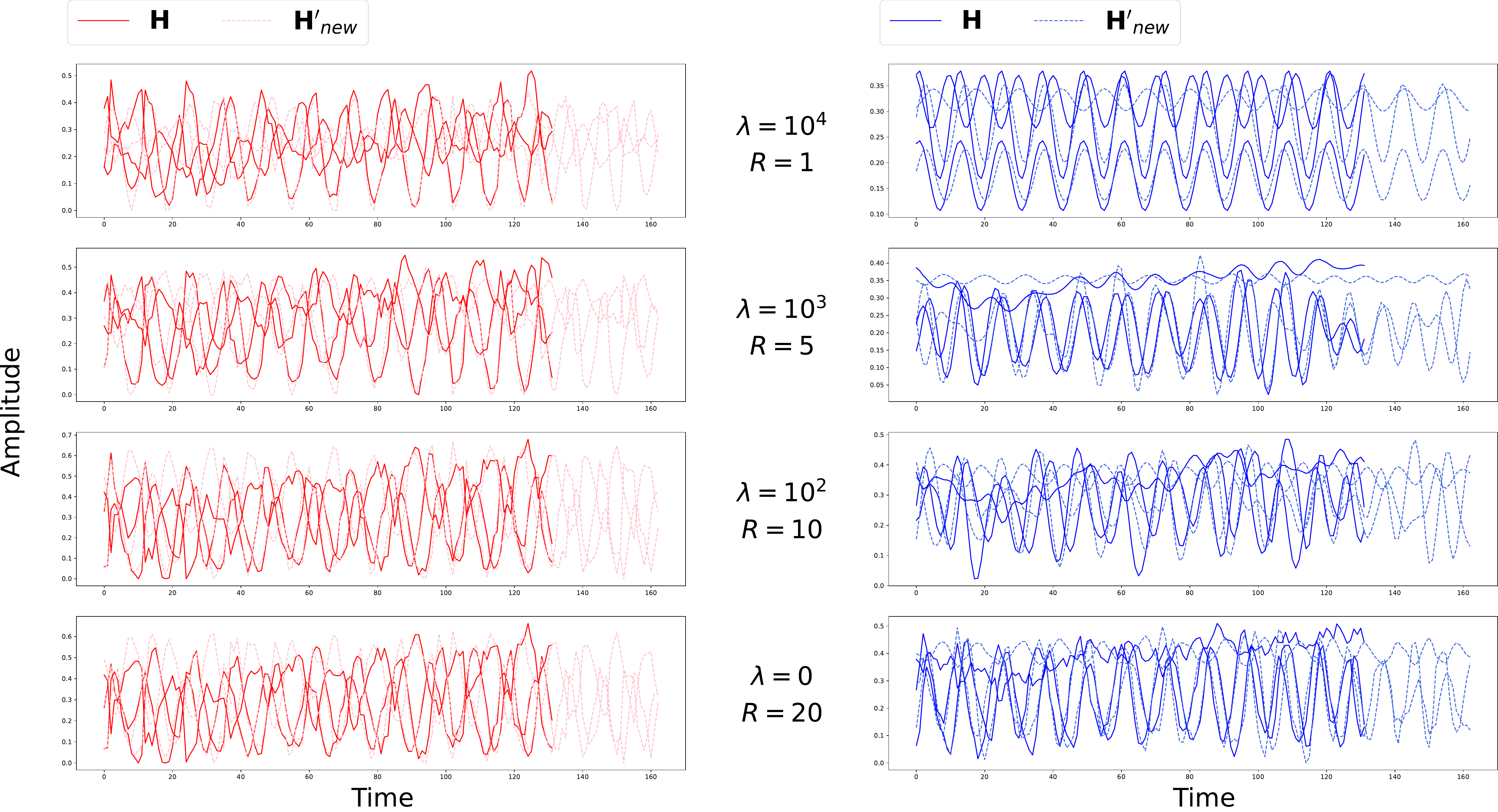}
\newline
    \newline
    \newline
\includegraphics[width=1 \linewidth]{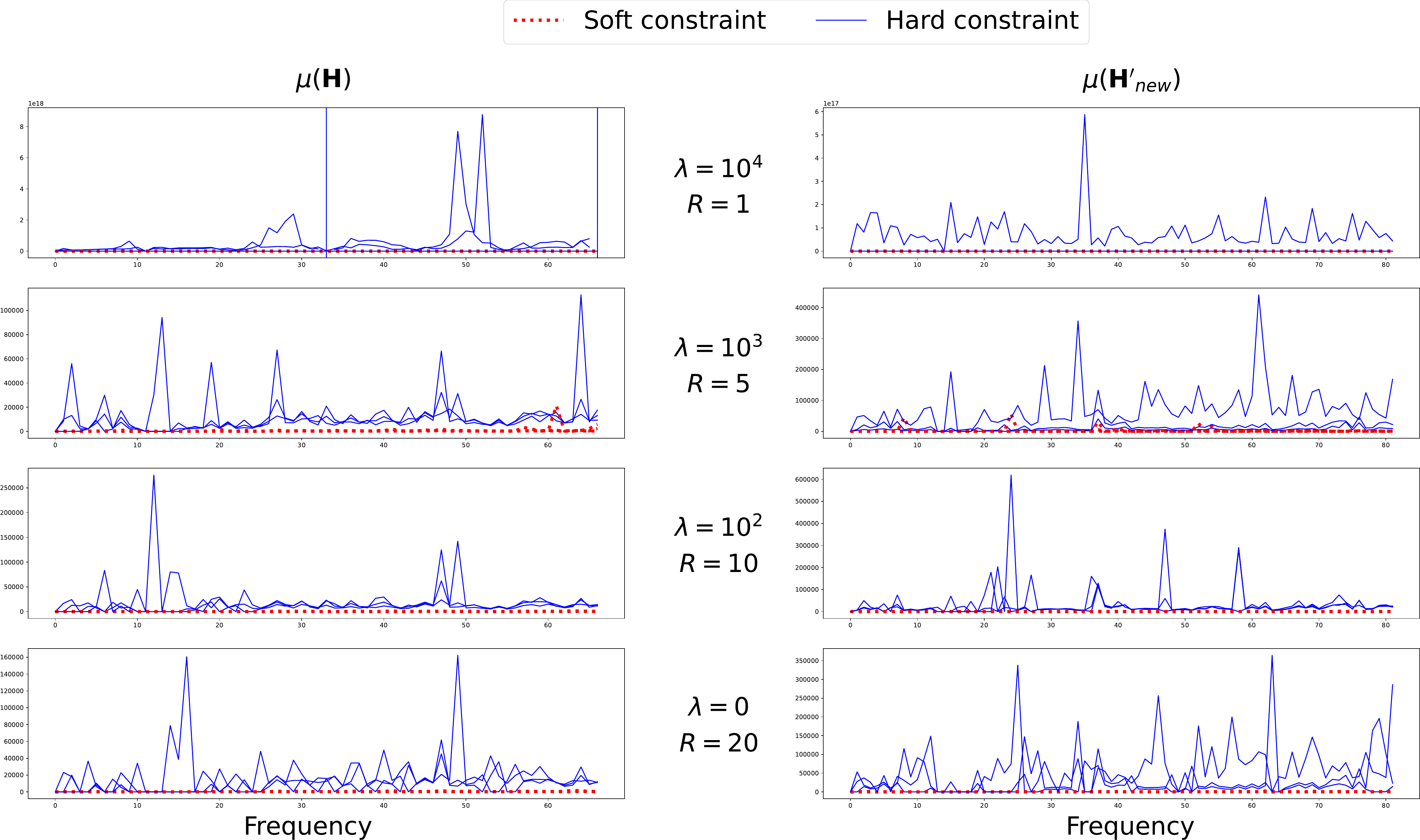}
    \caption{Yangtze data, $r=3$, $\xi=100$. Each row corresponds to regularization parameters $\lambda \in \left\{0, 10^2, 10^3, 10^4\right\}$ and remaining frequencies, specifically $R \in \left\{1, 5, 10, 20\right\}$. First row: spatial pattern of $\W$, Second row: temporal pattern of $\H$. Third row: $\mu(\H)$ and $\mu(\H'_{\textup{new}})$ respectively. This shows only positive frequencies (frequencies up to half of the length of the time-series) since they have the same value as the negative frequencies. Vertical lines in plots signify an infinite value.}
\label{fig:hardsoft}
\end{figure}

{\color{black} \subsection{Comparison among Lasso, soft and hard frequency regularization}

\begin{definition}[Inverse usage ratio of frequencies]
In Example \ref{ex:three norms}, we focused on reducing the amplitude of undominated frequencies. To measure the decrease in amplitude for the frequencies that are not dominant, 
we define the inverse usage ratio of frequencies as follows:
\begin{equation}
\label{def:Inverse usage ratio of frequencies}
\mu(\H) = \left[\left({\left| \widehat{\H}[s,k] \right| \over \sum\limits_{l} \left| \widehat{\H}[s,l] \right|}\right)^{-1}\right]_{\substack{0 \le s \le r-1, \\ 0 \le k \le T-1}} \in \R^{r \times T},
\end{equation}
where $\H \in \R^{r \times T}$. A large value of $\mu(\H)[s_0, k_0]$ indicates that the $k_0$th Fourier coefficient of the $s_0$th temporal pattern is small.
\end{definition}

\begin{example}
\label{ex:Lasso soft grace}
For the comparison of the Lasso and soft frequency regularization methods we consider the Yangtze dataset. Figure \ref{fig:soft-Lasso} shows that in soft frequency regularization, $\mu(\H)$ has larger values than Lasso. Similar to Example \ref{ex:three norms}, we observe that soft frequency regularization  reduces the usage of undominated (noise) frequencies more effectively than Lasso regularization, but it does not achieve the desired sparsity in the frequency domain.
\end{example}

\begin{figure}[!]
    \centering
    \includegraphics[width=1 \linewidth]{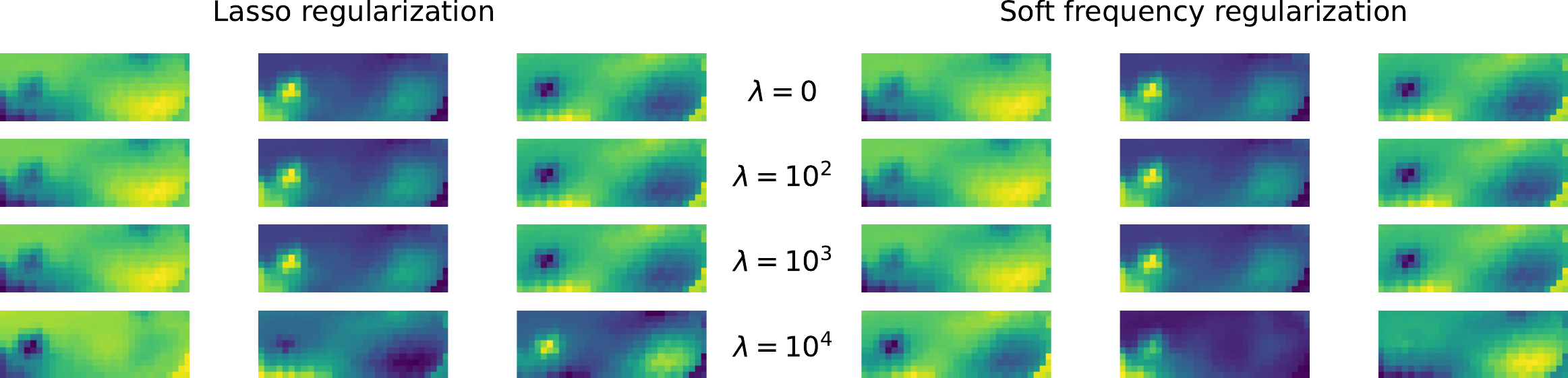}
    \newline
    \newline
    \newline
\includegraphics[width=1 \linewidth]{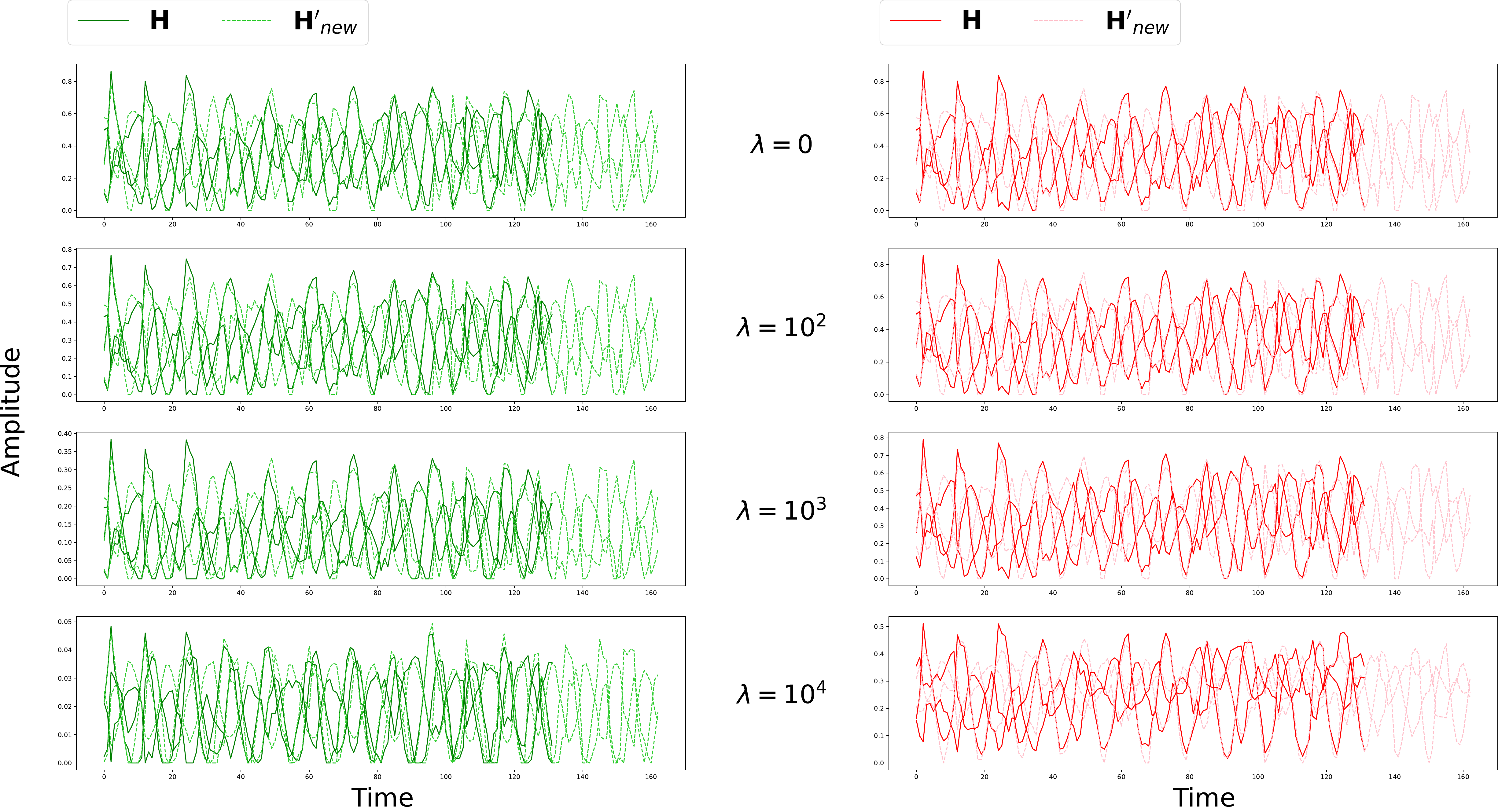}
\newline
    \newline
    \newline
\includegraphics[width=1 \linewidth]{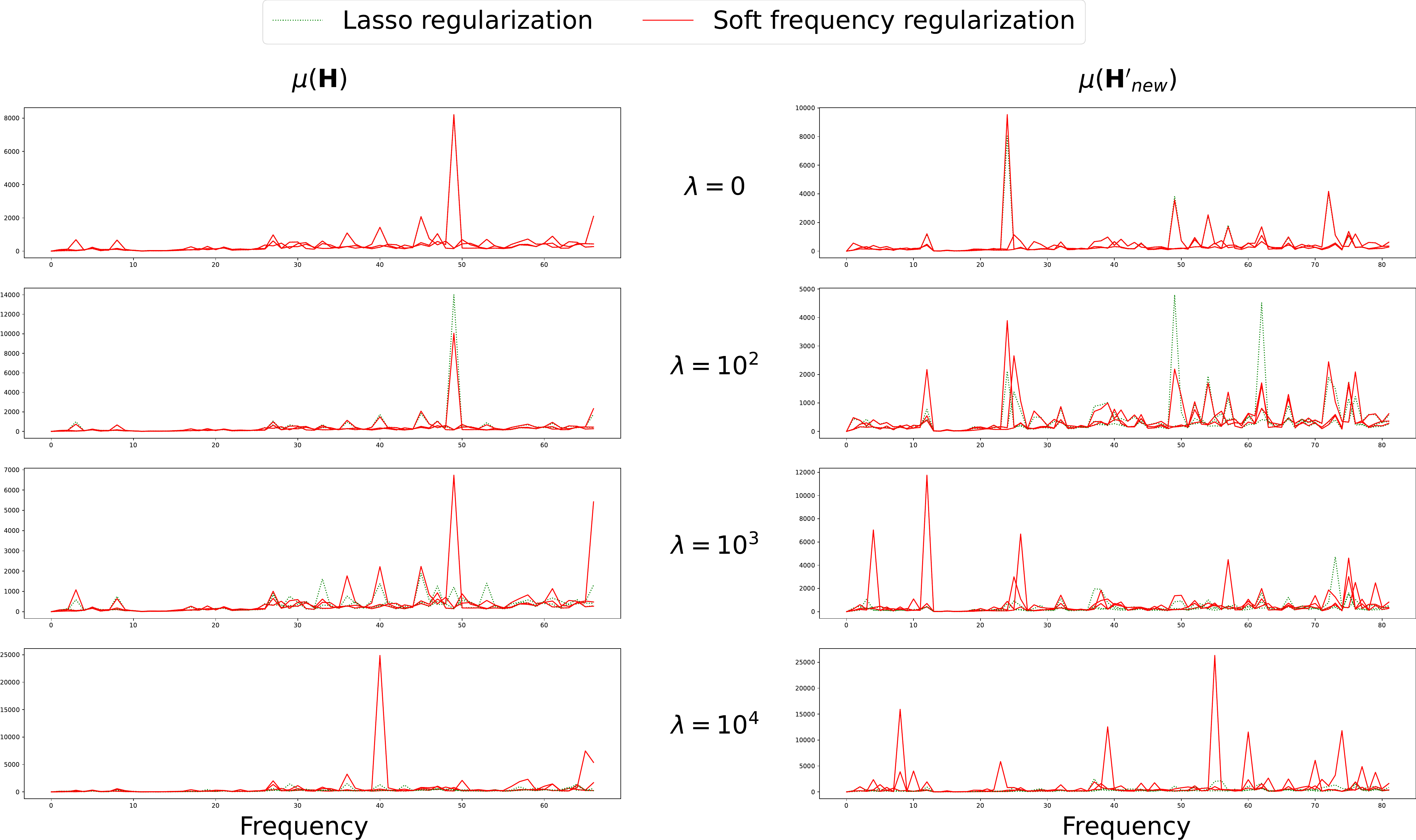}
\caption{Yangtze data: $r = 3$, $\xi = 100$. Each row corresponds to a choice of regularization parameters $\lambda \in \left\{0, 10^2, 10^3, 10^4\right\}$. First row: spatial pattern of $\W$, Second row: temporal pattern of $\H$. Third row: $\mu(\H)$ and $\mu(\H'_{\textup{new}})$, respectively. This shows only positive frequencies (frequencies up to half of the length of the time-series) since they have the same value as the negative frequencies.}
\label{fig:soft-Lasso}
\end{figure}

\begin{example}
\label{ex:soft vs hard}
In this example, we compare soft and hard frequency regularization methods for the Yangtze dataset. It can be observed that the elements of $\mu(\H)$ are higher with hard frequency regularization than soft frequency regularization. In this experiment, we can verify that hard frequency regularization is more effective than soft frequency regularization in minimizing the utilization of particular frequencies. 
Figure \ref{fig:hardsoft} verifies these results. Note that the reason why there is no always value of $\mu(\H)$ that diverges to $\infty$ with the hard constraint is explained in Algorithm \ref{alg:Heuristic}, where we prioritize the nonnegativity condition over the removal of specific frequencies in $\H$.
\end{example}}

\subsection{Accuracy of spatio-temporal prediction}

Now for the evaluation of the proposed method, consider the following  metric.
\begin{definition}[Nash–Sutcliffe efficiency (NSE), \cite{sun2020reconstruction}]
Nash–Sutcliffe efficiency (NSE) is defined as follows:
$$NSE = 1 -{\sum\limits_{i=0}^{T-1} \left(\ \overline{\X}_i - \overline{\X}_{rec,i} \right) \over \sum\limits_{i=0}^{T-1}  \left(\ \overline{\X}_i - \left\langle \overline{\X}  \right\rangle\right)},$$
where $\X_{rec}$ is the output of the model, $\overline{\X}$ is the spatial average of $\X$ and $\left\langle \cdot \right\rangle$ denotes the average in time. NSE ranges between $-\infty$ and $1$. If NSE is close to $1$, it is indicated that the model predicts perfectly. 
\end{definition}

We use the number of spatial-temporal patterns, denoted as $r$, as a hyperparameter, ranging from 2 to 20. Additionally, we chose additional parameters for $\xi \in \bigl \{10^{-3}, 10^{-2}, 10^{-1}, 1, 10, 10^2, 10^3, 10^4, 10^5 \bigr \}$, $\lambda \in \bigl\{10^{-1}, 1, 10,$ $10^2, 10^3, 10^4\bigr\}$, and $R \in \left\{10, 20, 30, 40 \right\} $. For each hyperparameter, we conduct 10 experiments and calculate the median value of the NSEs obtained from these 10 experiments. Table \ref{test:1} summarizes the result.

\begin{table}[H]
\centering
\includegraphics[width=1 \linewidth]{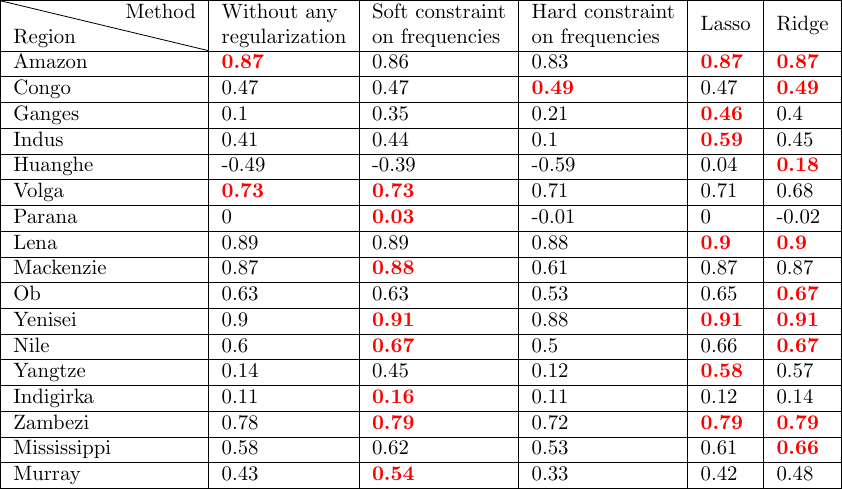}
\caption{Median value from 10 experiments. The numbers in boldface represent the highest NSE values in the methods.}
\label{test:1}
\end{table}

{\color{black} In these experimental results, we can observe that the soft frequency, Lasso, and Ridge regularization methods generally demonstrate the best predictive performance among all, as shown in Table \ref{test:1}. As confirmed in Example \ref{ex:soft vs hard}, hard frequency regularization is most effective in removing undominated frequencies compared to other regularization mehtods, but it appears less effective in predictive performance. 
This experiment shows that we can select each form of regularization depending on whether our analysis primarily centers on the time domain or the frequency domain. Our choice can also depend on our priority for interpretability and predictive accuracy.
}

\subsection{Atom removal}
\begin{figure}[!]
    \centering
    \includegraphics[width=1 \linewidth]{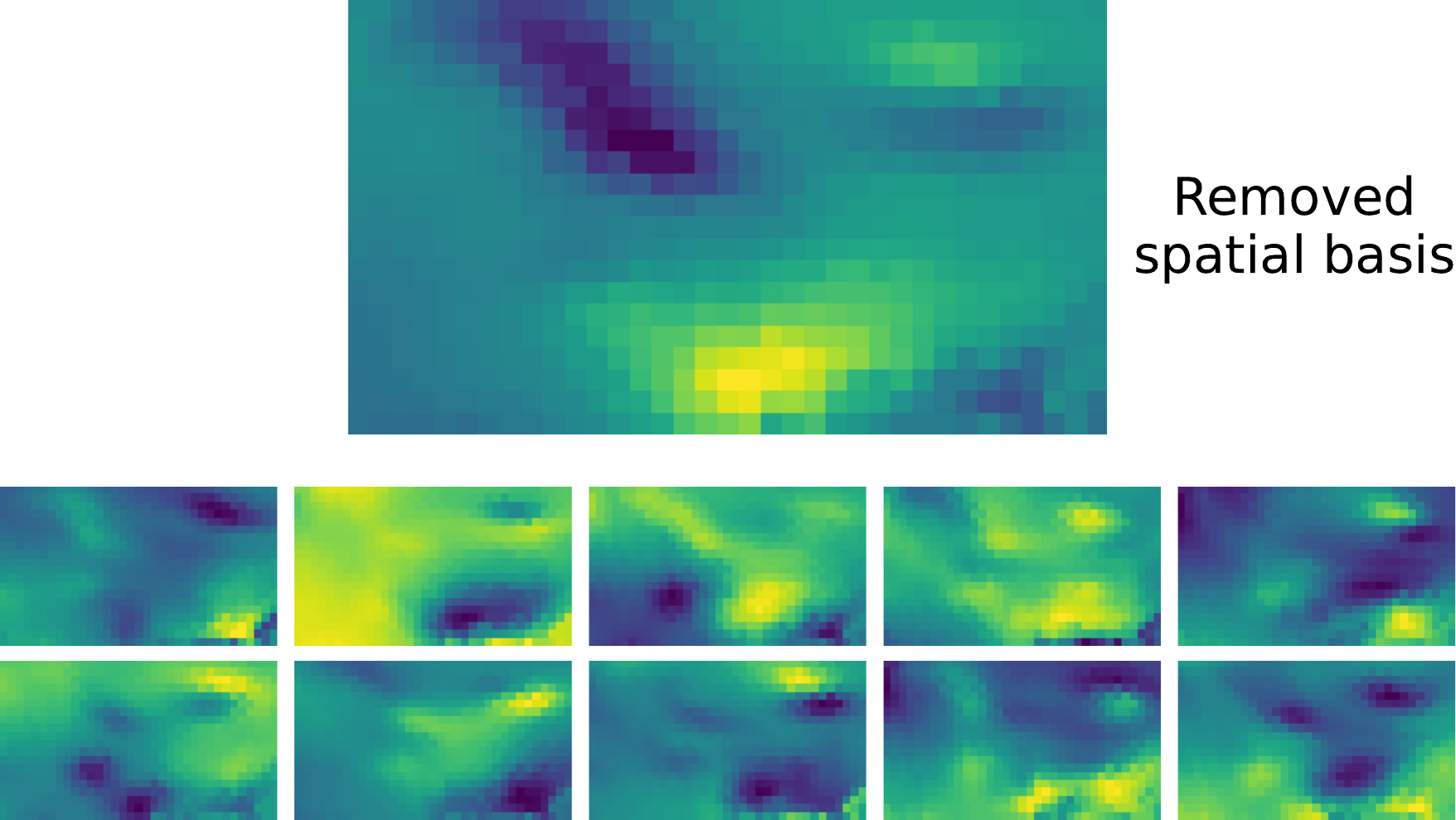}
    \newline
    \newline
    \newline
    \includegraphics[width=1 \linewidth]{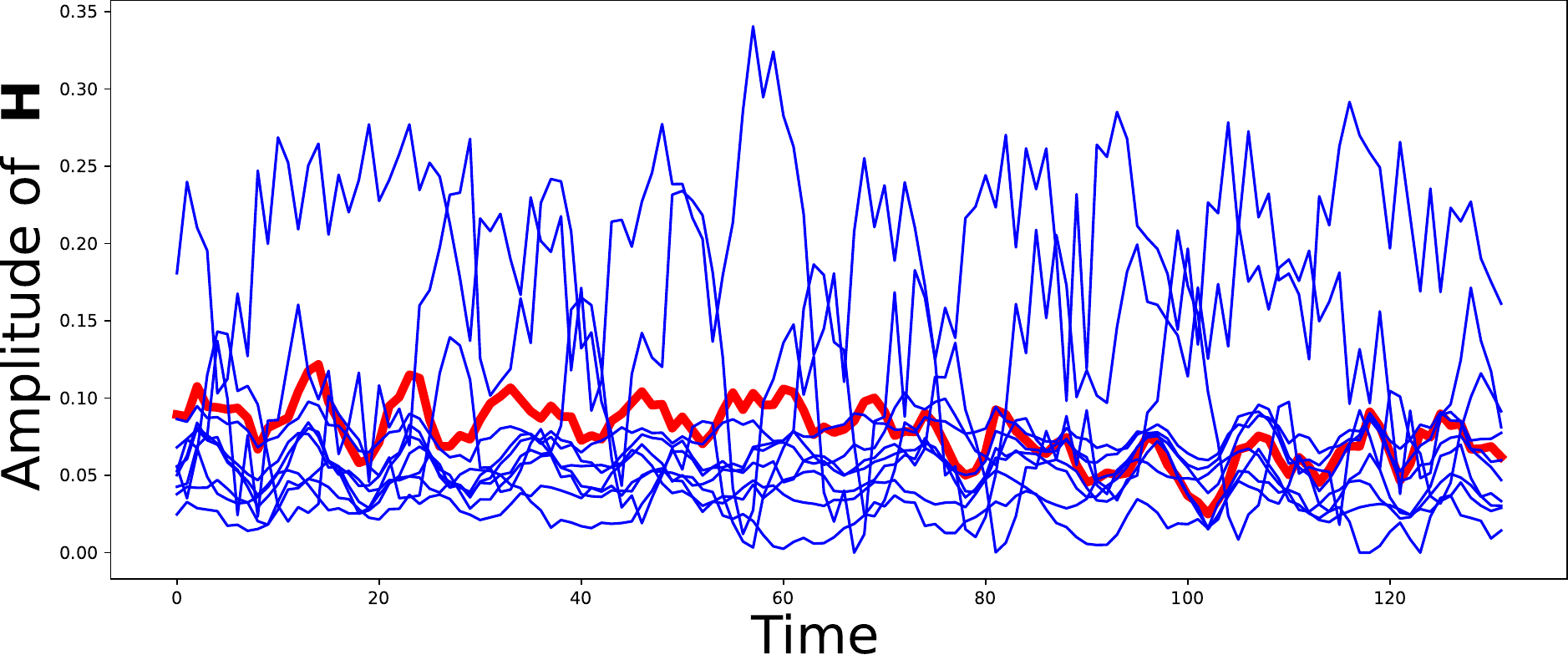}
    \newline
    \newline
    \includegraphics[width=1 \linewidth]{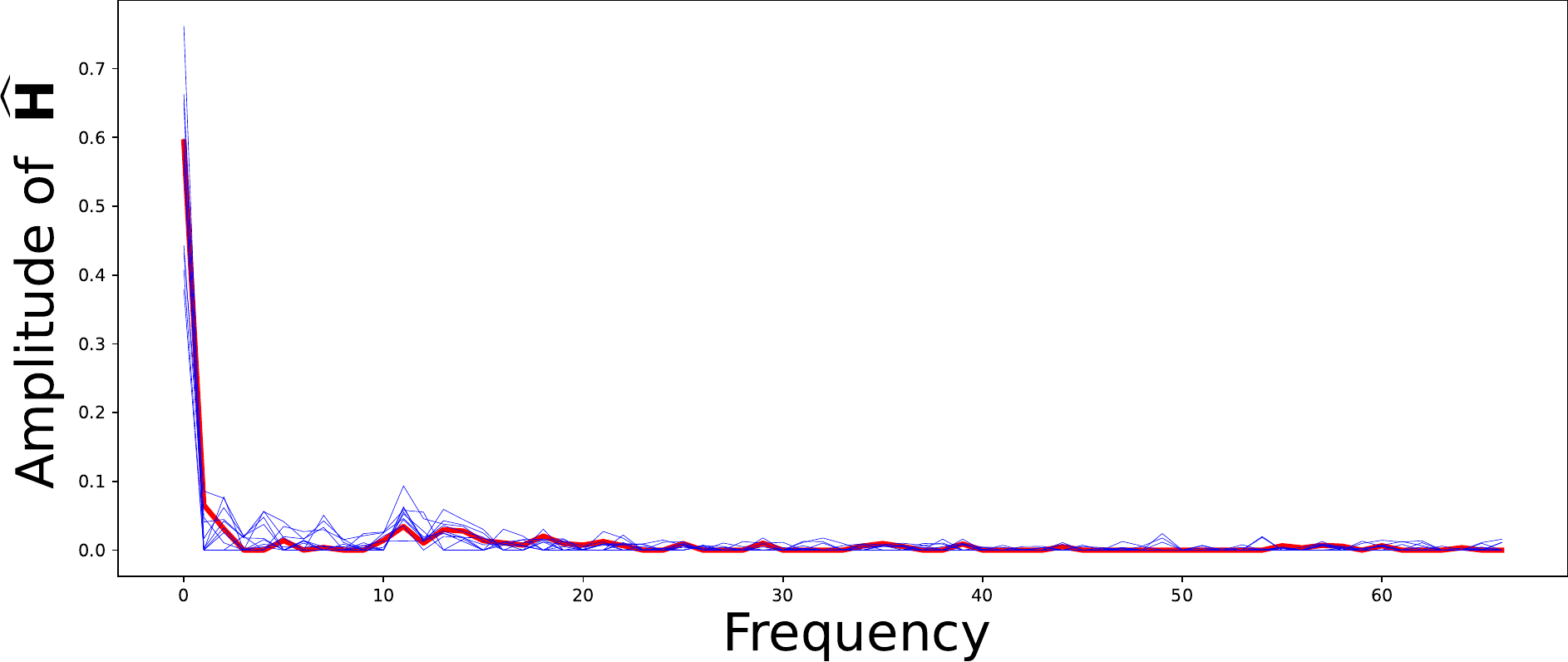}
    \caption{In the Mississippi region with 11 atoms, $\xi = 10^4$, and 30 remaining frequencies, removing one spatial and the corresponding temporal atom (bold red line in the middle and bottom figures) causes the NSE to change from 0.1 to 0.77. This suggests that the first atom contains noisy components.}
    \label{fig:atom_remove}
\end{figure}

The term `atom' refers to a spatial latent pattern of the given data. Note that some atoms could represent noise. Therefore, by removing such noisy atoms learned during the coding process, we may achieve better results. This process is similar to PCA, where we select the principal components that capture the most variation in the data. However, the key difference between PCA and the proposed method is the utilization of supervised information to identify the appropriate atoms. If the NSE value significantly increases after removing a particular atom, that atom can be considered as noise. Given the possibilities for discarding multiple atoms, we only consider removing a single atom. All hyperparameter settings are the same as those in above. As shown in Tables \ref{test:1}, \ref{tab:NSE_SUN_TABLE} and \ref{tab:atom_remove_table}, our method is comparable to previous research in the geophysical research community. The sharp increase in NSE in the Ob region is noteworthy. Prior to discarding the atoms under the hard constraint, the NSE value in the Ob region was 0.18. However, after discarding the atoms, it dramatically increased to 0.77. See Figure \ref{fig:atom_remove}.

\begin{table}[h]
    \centering
    \includegraphics[width=0.6 \linewidth]{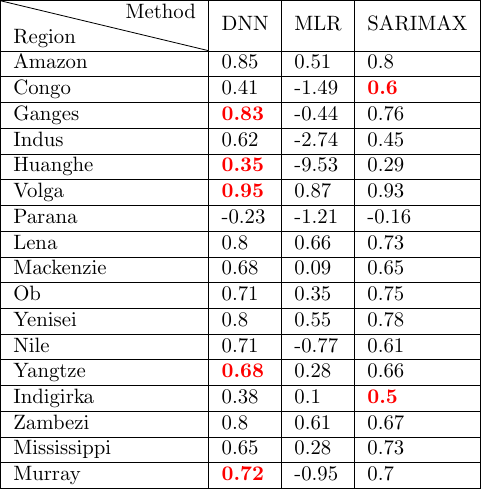}
    \caption{Median of NSE in previous research in the geophysical science \cite{sun2020reconstruction}. The boldface numbers indicate the highest NSE values in both the current table and Table \ref{tab:atom_remove_table}.}
    \label{tab:NSE_SUN_TABLE}
\end{table}

\begin{table}[!]
    \centering
    \includegraphics[angle=270,origin=c]{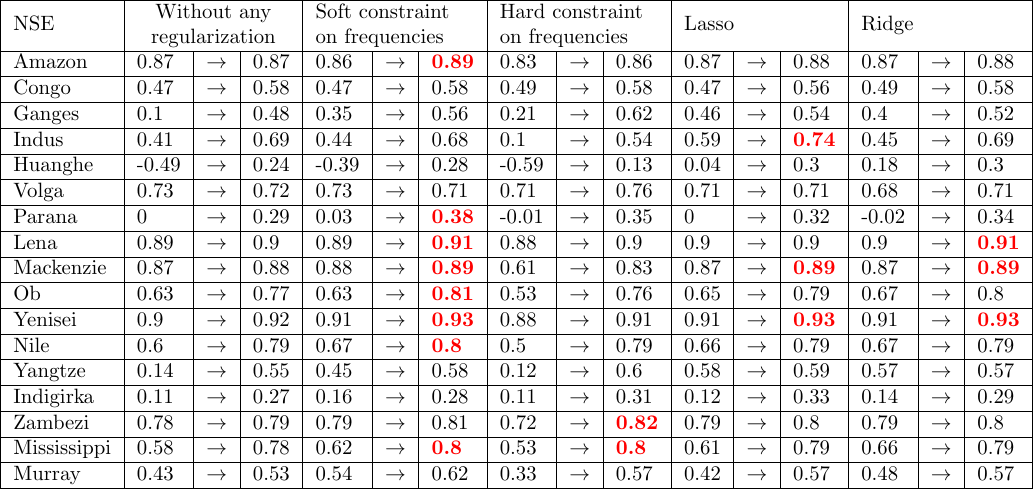}
    \caption{Median of NSE the final change is the best score before and after atom removal. The boldface numbers indicate the highest NSE values in both the current table and Table \ref{tab:NSE_SUN_TABLE}. Each arrow represents the change in NSE before and after discarding an atom.}
    \label{tab:atom_remove_table}
\end{table}

\section{Conclusion}
In this paper, we considered forecasting problems, specifically those arising in applications characterized by time periodicity, such as geophysical problems. In geophysical scenarios, spatial patterns exhibit changes with distinct periodicities, and frequency information becomes pivotal for accurate forecasting. To address this, we introduced novel methods based on Supervised low-rank Semi-Nonnegative Matrix Factorization (SSNMF), incorporating both soft and hard regularization in the frequency domain. These proposed methods aim to extract accurate temporal patterns from spatio-temporal data by leveraging essential periodicity information.

For the soft constraint approach, we proposed to use the Minkowski 1-norm for feature selection in the frequency domain, similar to Lasso. {\color{black} Although the amplitudes of undominated (noise) frequencies are reduced with soft regularization, those noisy frequencies are not completely eliminated.} Consequently, we introduced the hard constraint in the frequency domain, employing three operator splitting techniques. However, for the hard constraint it is required to provide prior knowledge about the frequencies to be removed. For this, we introduced the heuristic approach, which allows for the removal of specific frequencies in the frequency domain without requiring any prior knowledge. {\color{black} 
Consequently, the forecasting of spatio-temporal data through the soft frequency, Lasso, and Ridge regularization methods comparably align with prior research in geophysical science, while hard frequency regularization falls behind in this regard. Nonetheless, hard frequency regularization contributes to more concise interpretation by enforcing sparsity in the frequency domain. This affords us the flexibility to choose the most suitable regularization method, depending on whether our primary emphasis lies in the time domain or the frequency domain, as well as our priority between interpretability and predictive capacity. We  provided theoretical convergence results for the proposed methods. We also presented numerical results demonstrating the  effectiveness of the proposed methods using GRACE data from geophysical applications. These results show the benefits of taking frequency information into account for forecasting problems, enhancing both accuracy and interpretability.

As explained in this paper, the proposed method proves effective for problems characterized by time periodicity, particularly enhancing the interpretability of the provided data. However, there are instances when non-periodic events, such as anomalous occurrences, become significant in time-series data. Our current method might overlook such frequency information. In future work, we will attempt to further develop the proposed method to address situations where non-periodic characteristics also play a crucial role in the observed phenomena within the given data. Additionally, for this study, we did not conduct a comprehensive analysis of geophysical data, specifically GRACE data, which falls outside the scope of this paper. Instead, our emphasis lies on developing the methodology and providing proofs of performance.
In our future research, we will also attempt to apply the proposed method to domain-specific problems, such as geophysical problems, and perform a thorough analysis, emphasizing interpretability.


}

\section{Notation}

\begin{enumerate}
    \item $T, T_{tot}$ : Train time period and test period, p.\pageref{not:T,T_tot}.
    
    \item $\mathcal{X}, \mathcal{Y}_i \in \R^{A\times B \times T}$ : Main spatio-temporal data and supervision datas, p.\pageref{not:data tensor}.

    \item $r$ : The number of spatial/temporal patterns of $\mathcal{X}$, \pageref{not:number of patterns}.

    \item $\H$ : Temporal pattern on train period of $\mathcal{X}$, p.\pageref{not:temporal pattern}.
    
    \item $S$ : The number of data for supervision, p.\pageref{not:S}.

    \item $\H_{\textup{new}}'$ : Temporal pattern on train + test period of $\mathcal{X}$, p.\pageref{eq:inffer_H_new}.

    \item $\H_{\textup{new}}$ : Test period part in $\H_{\textup{new}}'$, p.\pageref{eq:inffer_H_new}.

    \item $\widetilde{\H}$ : Concatenation of $\H$ and $\H_{\textup{new}}$, p.\pageref{not:tilde H}.
    
    \item $\lVert \cdot \rVert_{F}$ : Frobenius norm, p.\pageref{eq:inffer_H_new}.

    \item  $\X, \Y_i \in \R^{d \times T}$ : Matricization of $\mathcal{X}$ and $\mathcal{Y}_i$, p.\pageref{not:matricization}.

    \item $\Y \in \R^{dS \times T}$ : Matrix obtained by stacking $\Y_1,\cdots, \Y_S$ vertically, p.\pageref{not:matricization}.

    \item $\mathcal{W}, \mathcal{W}',\W,\W'$ : Spatial pattern of $\mathcal{X}$, $\mathcal{Y}$ and its matricization, p.\pageref{not:spatial pattern}.
    
    \item $\widehat{\H}, \mathcal{F}_T$ : Fourier transform of $\H$ and Fourier transform matrix, p.\pageref{not:Fourier transform}.

    \item $\xi, \lambda$ : Regularization parameter of supervision data and frequency, p.\pageref{eq:loss function}.

    \item $\lVert \cdot \rVert_{1,M}$ : Minkowski 1-norm, p.\pageref{not:Minkowski norm}.

    \item $R$ : The number of remaining frequencies p.\pageref{eq:h}.

    \item ${\partial \over \partial z}, {\partial \over \partial \bar{z}}, \grad_{z}, \grad_{\bar{z}}$ : Wirtinger derivatives, p.\pageref{not:Wirtinger derivatives}.

    \item $\Re(z), \Im(z)$ : Real and imaginary part of $z$, p.\pageref{not:real,imaginary}.

    \item $\mu(\H)$ : Inverse usage ratio of frequencies of $\H$, p.\pageref{def:Inverse usage ratio of frequencies}
\end{enumerate}

\begin{acknowledgements}
We thank Jae-Seung Kim and Ki-Weon Seo at Seoul National University for their helpful discussions on the data used in this research.
\end{acknowledgements}

\noindent
{\small 
{\textbf{Funding}}
The second author  was partially supported by the NSF through grants DMS-2206296 and DMS-2010035.
The third author was supported by Samsung Electronics Co., Ltd (IO230407-05812-01), National Research Foundation of Korea (NRF) grant funded by the Korea government (MSIT) (No. RS-2023-00219980, No. 2022R1C1C1008491 and No. 2021R1A6A1A10042944), and POSCO HOLDINGS research fund (2022Q019). 
The fourth author was supported by the NRF  under the grant number 2021R1A2C3009648 and POSTECH Basic Science Research Institute under the NRF  2021R1A6A1A1004294412.}

\vskip .1in
\noindent
{\small 
{\textbf{Author Declaration and Conflict of Interest}}
The authors confirm that the manuscript has been read and approved by all named authors and that there are no other persons who satisfied the criteria for authorship but are not listed. The authors further confirm that the order of authors listed in the manuscript has been approved by all of us. The authors also  hereby certify that there is not any actual or potential conflict of interest.

\vskip .1in
\noindent
{\small 
{\textbf{Data availability}}
The datasets used in this work are publicly accessible through the following sources. The CSR GRACE/GRACE-FO RL06 data are obtained from \url{https://www2.csr.utexas.edu/grace/RL06_mascons.html}, ERA5 monthly averaged data on single levels data from \url{https://cds.climate.copernicus.eu/cdsapp#!/dataset/reanalysis-era5-single-levels-monthly-means?tab=overview},  GLDAS Noah Land Surface Model data from \url{https://disc.gsfc.nasa.gov/datasets/GLDAS_NOAH025_M_2.1/summary?keywords=gldas}, and the latitude and longitude information for the river basin  from \url{https://hydro.iis.u-tokyo.ac.jp/~taikan/TRIPDATA/Data/rivers.idx}. The data availability of the proposed methods should be directed to the authors. 
}

\bibliographystyle{spmpsci}
\bibliography{bib.bib}

\end{document}